\documentclass[conference]{IEEEtran}
\IEEEoverridecommandlockouts

\usepackage{cite}
\usepackage{amsmath,amssymb,amsfonts}
\usepackage{mathtools}
\usepackage{amsthm}
\usepackage{graphicx}
\usepackage{textcomp}
\usepackage[table]{xcolor}
\usepackage{booktabs}
\usepackage{multirow}
\usepackage{makecell}
\usepackage{array}
\usepackage{tabulary}
\usepackage{colortbl}
\usepackage{subcaption}
\usepackage{balance}
\usepackage{url}
\usepackage[hidelinks]{hyperref}
\usepackage[capitalize,noabbrev]{cleveref}

\definecolor{softred}{RGB}{205, 92, 92}
\definecolor{paperblue}{RGB}{100, 149, 237}
\definecolor{lightgray}{HTML}{DDDDDD}

\makeatletter
\newcommand{\ie}{\emph{i.e.}\@ifnextchar.{\!\@gobble}{}}
\newcommand{\eg}{\emph{e.g.}\@ifnextchar.{\!\@gobble}{}}
\newcommand{\etc}{etc\@ifnextchar.{}{.\@}}
\makeatother

\def\BibTeX{{\rm B\kern-.05em{\sc i\kern-.025em b}\kern-.08em
    T\kern-.1667em\lower.7ex\hbox{E}\kern-.125emX}}
\makeatletter
\renewcommand{\IEEEauthorrefmark}[1]{\textsuperscript{#1}}
\makeatother
\begin{document}

\title{Towards Multi-Source Domain Generalization for Sleep Staging with Noisy Labels}

\author{
\IEEEauthorblockN{
Kening Wang\IEEEauthorrefmark{1},
Di Wen\IEEEauthorrefmark{1},
Yufan Chen\IEEEauthorrefmark{1},
Ruiping Liu\IEEEauthorrefmark{1},
Junwei Zheng\IEEEauthorrefmark{1,2}
}
\IEEEauthorblockN{
Jiale Wei\IEEEauthorrefmark{1},
Kailun Yang\IEEEauthorrefmark{3},
Rainer Stiefelhagen\IEEEauthorrefmark{1},
and Kunyu Peng\IEEEauthorrefmark{1}\textsuperscript{,*}
}
\IEEEauthorblockA{\IEEEauthorrefmark{1}Karlsruhe Institute of Technology, Germany}
\IEEEauthorblockA{\IEEEauthorrefmark{2}ETH Zurich, Switzerland}
\IEEEauthorblockA{\IEEEauthorrefmark{3}Hunan University, China}
\IEEEauthorblockA{
\textsuperscript{*}Corresponding author:
Kunyu Peng (Email: kunyu.peng@kit.edu)
}
}
\maketitle

\begin{abstract}
Automatic sleep staging is a multimodal learning problem involving heterogeneous physiological signals such as EEG and EOG, which often suffer from domain shifts across institutions, devices, and populations. In practice, these data are also affected by noisy annotations, yet label-noise-robust multi-source domain generalization remains underexplored. We present the first benchmark for Noisy Labels in Multi-Source Domain-Generalized Sleep Staging (NL-DGSS) and show that existing noisy-label learning methods degrade substantially when domain shifts and label noise coexist. To address this challenge, we propose \textbf{FF-TRUST}, a domain-invariant multimodal sleep staging framework with Joint Time-Frequency Early Learning Regularization (JTF-ELR). By jointly exploiting temporal and spectral consistency together with confidence-diversity regularization, \textbf{FF-TRUST} improves robustness under noisy supervision. Experiments on five public datasets demonstrate consistent state-of-the-art performance under diverse symmetric and asymmetric noise settings.

\end{abstract}


\section{Introduction}
Sleep is fundamental to physiological homeostasis and human health, while insufficient or poor-quality sleep is strongly associated with cognitive impairment and chronic diseases~\cite{czeisler2015duration,sharma2010sleep,neckelmann2007chronic,baglioni2011clinical}. Automatic sleep staging from polysomnography (PSG) is therefore a critical problem in biomedical signal analysis, requiring robust interpretation of heterogeneous and temporally structured modalities such as electroencephalogram (EEG) and electrooculogram (EOG). According to the American Academy of Sleep Medicine (AASM), sleep is scored into wake (W), non-rapid eye movement (NREM; N1--N3), and rapid eye movement (REM) stages~\cite{berry2012aasm}. However, manual scoring is labor-intensive and subject to inter-rater variability~\cite{aboalayon2016sleep}, motivating reliable deep sequence-to-sequence models for automated sleep staging.

Recent deep learning methods have achieved promising performance on multimodal sleep staging~\cite{ma2024sleepmg,ma2024exploring,jo2025measure,lv2022causality}. Nevertheless, most studies assume clean annotations and relatively controlled data distributions, which contrasts sharply with real-world PSG collected across institutions, devices, and populations. Such data naturally suffer from substantial domain shifts and annotation noise. To address cross-dataset variability, Domain Generalization of Sleep Staging (DGSS) has recently emerged as an important setting~\cite{wang2024generalizable}. In particular, SleepDG~\cite{wang2024generalizable} studies multi-source cross-dataset DGSS, a realistic yet challenging scenario involving inter-institute, inter-subject, and inter-device heterogeneity. However, the impact of noisy labels in multi-source DGSS remains largely unexplored.

This omission is critical. Existing multi-source domain generalization methods~\cite{wang2024generalizable,hu2022domain,wei2024multi} typically assume clean supervision, although clinical annotations are costly, subjective, and error-prone. In multi-source DGSS, label noise can amplify distributional discrepancies and severely degrade target-domain generalization, as also observed in visual recognition~\cite{peng2024advancing}. For multimodal physiological sequences, the challenge is even more pronounced: noisy labels and domain shifts jointly corrupt temporal dynamics and spectral patterns, which are essential for reliable sleep-stage representation learning. Robust learning under noisy supervision is therefore indispensable for trustworthy and deployable sleep staging systems.

To bridge this gap, we introduce the first benchmark for \emph{Noisy Labels in multi-source Domain-Generalized Sleep Staging} (NL-DGSS). Built upon SleepDG~\cite{wang2024generalizable}, the NL-DGSS benchmark adapts five representative noisy-label learning methods, \ie, NPN~\cite{sheng2024adaptive}, JAL~\cite{wang2025joint}, DISC~\cite{li2023disc}, ELR~\cite{liu2020early}, and FedDSHAR~\cite{lin2025feddshar}, to cross-dataset multimodal sleep staging. Our analysis shows that methods originally designed for \emph{i.i.d.} visual data degrade substantially under heterogeneous physiological domain shifts, revealing a mismatch between visual-space noise-robust learning and multimodal biomedical sequence modeling.

We further propose \textbf{FF-TRUST}, a domain-invariant multimodal sleep staging framework with a novel \textbf{J}oint \textbf{T}ime-\textbf{F}requency \textbf{E}arly \textbf{L}earning \textbf{R}egularization (\textbf{JTF-ELR}). \textbf{FF-TRUST} inherits the sequence-to-sequence autoencoding and sequence-/epoch-level alignment of SleepDG~\cite{wang2024generalizable}, while explicitly improving robustness to noisy supervision. Specifically, \textbf{JTF-ELR} enforces EMA-based historical embedding consistency in both temporal and spectral domains through \textbf{TimeELR} and \textbf{FourierELR}. In addition, a novel forward-free confidence diversity regularization (\textbf{FF-CDR}) avoids explicit noise-transition estimation, encourages confident yet balanced predictions, and mitigates class collapse under severe label noise.

Extensive experiments on five public sleep staging datasets demonstrate that \textbf{FF-TRUST} consistently achieves state-of-the-art performance on the NL-DGSS benchmark. Under severe symmetric noise ($0.6$), \textbf{FF-TRUST} outperforms the best baseline by $1.69\%$ ACC and $3.00\%$ MF1 on average across target domains, highlighting the effectiveness of time-frequency consistency and forward-free regularization for robust domain-generalizable sleep staging.

\begin{itemize}
    \item We present the first systematic study of \emph{Noisy Labels in Domain-Generalized Sleep Staging} (NL-DGSS) and establish a benchmark by adapting $5$ representative noisy-label learning methods to multi-source DGSS.

    \item We propose \textbf{FF-TRUST}, the first NL-DGSS framework that combines domain-invariant multimodal sleep staging with label-noise-robust time--frequency early learning regularization on sequence-level latent representations.

    \item We introduce \textbf{JTF-ELR}, consisting of \textbf{TimeELR}, \textbf{FourierELR}, and forward-free confidence diversity regularization, to exploit temporal-spectral stability while avoiding fragile noise-transition estimation.

    \item Experiments on $5$ public sleep staging datasets show that \textbf{FF-TRUST} consistently outperforms strong baselines under diverse noise settings, achieving state-of-the-art performance in NL-DGSS.
\end{itemize}

\section{Related Work}
\noindent\textbf{Domain Generalization in Automatic Sleep Staging.} Automatic sleep staging is naturally formulated as a sequence prediction problem with strong temporal dependence and spectral structure. While deep models from raw EEG match or outperform hand-crafted pipelines \cite{Supratak2017DeepSleepNetAM, Supratak2020TinySleepNetAE}, real-world deployment requires \emph{Domain Generalization} (DG) across heterogeneous cohorts, hardware, and channel configurations \cite{gulrajani2021in, zhou2021domain_generalization_survey}. In this context, \emph{Domain Generalization of Sleep Staging} (DGSS) has evolved from robust architectural designs like U-Sleep \cite{perslev2021u} to algorithmic invariance via minimal-sufficient representations \cite{jo2025measure}, and multi-modal modeling \cite{ma2024sleepmg}. Crucially, cross-dataset evaluation must also account for the intrinsic ambiguity of sleep labels and rater disagreement \cite{rossi2025sleepyland}. However, current DGSS approaches typically assume reliable source supervision; when label noise is dataset-dependent and co-occurs with domain shift, invariance objectives can be systematically misdirected, a vulnerability also noted in other noisy DG regimes in the visual domain~\cite{peng2024mitigating, peng2025erelifm}.

\noindent\textbf{Learning with Label Uncertainty and Noise.} 
Label uncertainty is intrinsic to sleep staging, especially in transitional stages such as N1 where inter-rater agreement is low~\cite{lee2022interrater}. Existing studies address this through multi-expert modeling~\cite{van2023modeling}, while Label Noise Learning (LNL) methods explore sample selection~\cite{wei2020combating}, semi-supervised refinement~\cite{li2020dividemix}, and robust regularization~\cite{liu2020early}. However, transition-matrix correction~\cite{patrini2017making} is brittle under domain shift, and temporal dependencies may amplify early labeling errors~\cite{nagaraj2024learning}. For EEG-based sleep staging, stable time-frequency patterns provide more reliable supervision than noise-sensitive raw waveforms. Motivated by this, we study \emph{Noisy-Label Domain-Generalized Sleep Staging (NL-DGSS)} and propose \textbf{FF-TRUST}, which leverages EMA-based time-frequency consistency (\textbf{JTF-ELR}) and a forward-free objective to mitigate class-dependent corruption without fragile transition estimation under shift.

\section{Benchmark}
\subsection{Label Noise Settings}

In real-world sleep staging, annotation noise is difficult to collect and quantify reliably, since clean ground-truth labels are typically unavailable and inter-rater disagreement varies across datasets, scorers, and clinical protocols. Therefore, following common practice in label-noise learning, we adopt controlled synthetic corruption protocols to simulate symmetric and asymmetric noise. Although such a simulation does not perfectly reproduce real clinical annotation uncertainty, it provides a reproducible and standardized benchmark for evaluating robustness under different noise levels~\cite{li2023disc,sheng2024adaptive}. Our NL-DGSS benchmark considers two types of label noises, \ie, symmetric label noise and asymmetric label noise under various label noise ratios according to \cite{sheng2024adaptive}.

\noindent\textbf{Symmetric Label Noise.}
Following previous work~\cite{ma2020normalized}, we adopt symmetric label noise, where each clean label is independently corrupted with probability $\eta$ and replaced by a uniformly sampled incorrect class.
This setting simulates random annotation errors without introducing class-dependent bias.
In our experiments, we consider three noise rates, i.e., $\eta\in\{0.2,0.4,0.6\}$.

\noindent\textbf{Asymmetric Label Noise.}
To simulate structured annotation errors in sleep staging, we adopt a \emph{clinically informed synthetic} class-dependent noise model rather than uniform random corruption. Specifically, for each clean label $\mathbf{y}$, we flip the label with probability $\eta$; if a corruption occurs, the noisy label $\mathbf{\tilde{y}}$ is sampled from a fixed transition matrix $\mathbf{T} \in \mathbb{R}^{C \times C}$, where $\mathbf{T}_{ij} = P(\mathbf{\tilde{y}}=j \mid \mathbf{y}=i)$ and $\sum_{j=1}^{C} \mathbf{T}_{ij} = 1$. In contrast to symmetric noise, the off-diagonal entries of $\mathbf{T}$ are non-uniform: transitions between clinically plausible or commonly confused stages are assigned larger probability mass, such as $\mathrm{N1}$ and $\mathrm{N2}$, as well as
$\mathrm{N2}$ and $\mathrm{N3}$ while implausible transitions receive a very low probability. Importantly, the transition probabilities are not necessarily
symmetric, i.e., $T_{ij} \neq T_{ji}$.. This design is motivated by the fact that sleep-scoring disagreement is more likely to occur near ambiguous stage boundaries than uniformly across all classes. We emphasize that this asymmetric setting remains a \emph{synthetic approximation} of structured annotation uncertainty, intended to provide a controlled benchmark for evaluating robustness to class-dependent label corruption under domain shift, rather than a direct substitute for real inter-rater disagreement.

\begin{figure*}[t]
    \centering
    \includegraphics[width=\linewidth, trim=0 14.3cm 0 0, clip]{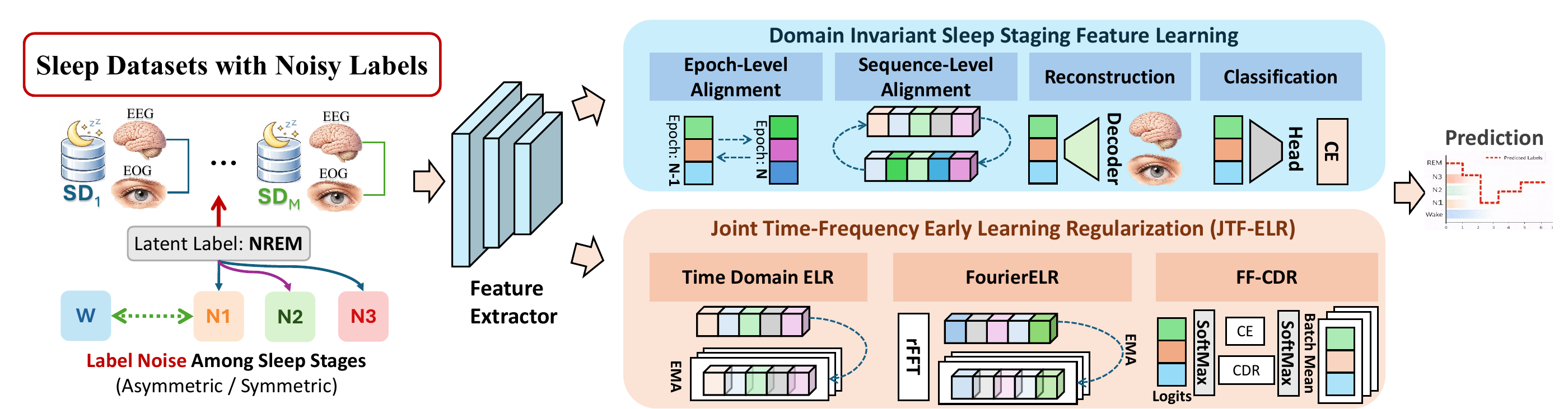}
    \caption{An overview of \textbf{FF-TRUST} architecture, comprising domain-invariant sleep staging feature learning and the proposed joint time--frequency early learning regularization. CE denotes cross-entropy loss, and CDR denotes confidence diversity regularization.}
    \vskip-2ex
    \label{fig:main}
\end{figure*}

\subsection{Baselines}
All experiments were conducted within the SleepDG framework~\cite{wang2024generalizable}, which provides a unified pipeline for domain generalization in sleep staging. We compared the proposed method with representative baselines from major noisy-label learning paradigms, which include:

\noindent\textbf{Instance-level selection and correction methods}, \eg, \textit{DISC}~\cite{li2023disc}, which could identify clean samples based on learning dynamics.

\noindent\textbf{Asymmetric-noise-aware approaches}, \eg, \textit{JAL}~\cite{wang2025joint}, which explicitly model class-dependent label transitions, \textit{NPN}~\cite{sheng2024adaptive} that focuses label uncertainty via partial and negative learning without prior noise, and \textit{ELR}~\cite{liu2020early}, which regularizes probability to avoid overmemorization.

\noindent\textbf{Robust time-series frameworks}, \eg,  \textit{FedDSHAR}~\cite{lin2025feddshar} adopts training on clean and noisy subsets.

\section{Methodology}
\subsection{Domain Invariant Sleep Staging}
\label{sec:4.1}
In this subsection, we introduce the preliminaries regarding domain invariant sleep staging representation learning, which serves as the foundation of our \textbf{FF-TRUST}, following~\cite{wang2024generalizable}. Our method is described in Figure~\ref{fig:main}.
Generalizable sleep staging is formulated as a multi-source domain generalization problem, where each sleep dataset is treated as a distinct domain. Given multiple labeled source domains $S=\{\mathcal{D}_S^i=(X_S^i,Y_S^i)\}_{i=0}^{M-1}$, where $i \in [0,..., M-1]$ denotes the index of the source domain, and $X_S^i$ and $Y_S^i$ indicate set of sequential data and labels, the goal is to learn a model that generalizes well to an unseen target domain.

Following prior work~\cite{wang2024generalizable}, the prediction function is decomposed as $h=f\circ g$, where $g$ denotes a feature encoder (consisting of CNN layers for basic representation learning and Transformer block for temporal aggregation) and $f$ is a classifier. The learning objective is shown in Eq.~\ref{eq:3}.
\begin{equation}
\label{eq:3}
\min_{f,g}\; \mathbb{E}_{(\mathbf{x},\mathbf{y})\sim P(X^S,Y^S)} \mathcal{L}(f(g(\mathbf{x})),\mathbf{y}) + \lambda\,\mathcal{L}_{\mathrm{reg}},
\end{equation}
where $\mathcal{L}_{\mathrm{reg}}$ promotes domain-invariant representations.

SleepDG~\cite{wang2024generalizable} instantiates $\mathcal{L}_{\mathrm{reg}}$ through four complementary loss terms that jointly serve as the base training objective, including classification, reconstruction, and feature alignment at both epoch and sequence levels, as shown in Figure~\ref{fig:main}.
A sleep recording is represented as a sequence of $N_e$ epochs, where $\mathbf{x}_k$ and $\hat{\mathbf{x}}_k$ denote the input signal of the $k$-th epoch and its reconstruction.
The classifier predicts $C$ sleep stages for each epoch.
$\mathbf{F}_i$ denotes the set of epoch-level features extracted from source domain $i$, and $\mathbf{R}_i$ denotes the expected inter-epoch correlation matrix computed from sequence features of domain $i$.

\noindent\textbf{Classification Loss.}
Sleep stage prediction is supervised using a cross-entropy loss applied at the epoch level, as shown in Eq.~\ref{eq:4}, where $N_e$ denotes the number of epochs.
\begin{equation}
\label{eq:4}
\mathcal{L}_{\mathrm{cls}}
= - \sum_{k=0}^{N_e-1} \sum_{c=0}^{C-1}
\mathbf{y}_{k,c} \log \hat{\mathbf{y}}_{k,c}.
\end{equation}

\noindent\textbf{Reconstruction Loss.}
To encourage informative and stable representations across domains, an autoencoder structure is utilized and the reconstruction error is minimized, as shown in Eq.~\ref{eq:5}.
\begin{equation}
\label{eq:5}
\mathcal{L}_{\mathrm{rec}}
= \frac{1}{N_e} \sum_{k=0}^{N_e-1}
\| \mathbf{x}_k - \hat{\mathbf{x}}_k \|_2^2 .
\end{equation}

\noindent\textbf{Epoch-Level Feature Alignment.}
To reduce domain shift at the epoch level, marginal feature distributions are aligned across source domains
by matching first-order and second-order statistics of the learned representations, as shown in Eq.~\ref{eq:6}.
\begin{equation}
\label{eq:6}
\mathcal{L}_{\mathrm{ep}}
= \sum_{i \neq j}
\Big(
\| \mathbb{E}(\mathbf{F}_i) - \mathbb{E}(\mathbf{F}_j) \|_2^2
+ \| \mathrm{Cov}(\mathbf{F}_i) - \mathrm{Cov}(\mathbf{F}_j) \|_F^2
\Big),
\end{equation}

where $(i, j) \in S$, $\mathrm{Cov}$ indicates covariance matrix, and $\| \cdot \|_F^2$ denotes the squared matrix Frobenius norm.

\noindent\textbf{Sequence-Level Feature Alignment.}
Beyond individual epochs, temporal dependencies are aligned across sleep stages by matching inter-epoch correlation structures.
Specifically, let $\mathbf{R}_i$ denote the expected Pearson correlation matrix
computed from sequence features in source domain $i \in \left[0, M-1 \right]$.
The sequence-level alignment loss is defined as Eq.~\ref{eq:7}.
\begin{equation}
\label{eq:7}
\mathcal{L}_{\mathrm{seq}} = \sum_{i \neq j} \| \mathbf{R}_i - \mathbf{R}_j \|_F^2.
\end{equation}

Collectively, these losses form the base training objective of domain invariant feature learning, which is shown as Eq.~\ref{eq:8}.
\begin{equation}
\label{eq:8}
\mathcal{L}
=
\mathcal{L}_{\mathrm{cls}}
+ \lambda_1 \mathcal{L}_{\mathrm{rec}}
+ \lambda_2 \mathcal{L}_{\mathrm{ep}}
+ \lambda_3 \mathcal{L}_{\mathrm{seq}},
\end{equation}
where $\lambda_1$, $\lambda_2$, and $\lambda_3$ are trade-off coefficients.

\subsection{Joint Time-Frequency Early Learning Regularization (JTF-ELR)}
\label{sec:4.2}
While SleepDG~\cite{wang2024generalizable} provides a strong domain-generalization baseline, its original objective
assumes reliable supervision.
In this work, we focus on the more challenging setting of noisy labels and introduce \textbf{J}oint \textbf{T}ime-\textbf{F}requency \textbf{E}arly \textbf{L}earning \textbf{R}egularization (\textbf{JTF-ELR}), which enhances the original SleepDG~\cite{wang2024generalizable} objective with three additional
regularization components.
Importantly, all domain invariant feature learning terms are preserved.

\noindent\textbf{Early-Learning Regularization (ELR) in Time Domain.}
Neural networks tend to fit clean labels in early training stages before memorizing noisy labels, which is verified on the LNL problem for visual space tasks~\cite{liu2020early}.
To exploit this property, we first incorporate Early-Learning Regularization (ELR) to stabilize the prediction of the time domain. Let $\mathbf{z} \in \mathbb{R}^{B \times T \times C}$ denote the logits, where $B$ is the batch size, $T$ is the sequence length, and $C$ is the number of classes.
We reshape the logits into token-level representations and denote by
$\mathbf{z}_n \in \mathbb{R}^{C}$ the logits corresponding to token $n$, where $n=1,\ldots,B*T$ and each token corresponds to a $(b,t)$ position in the original sequence. The predicted class probability vector is defined as $\mathbf{p}_n = \mathrm{SoftMax}(\mathbf{z}_n)$.
For each training epoch and for prediction token $n$, an Exponential Moving Average (EMA) buffer $\mathbf{P}_n\in\mathbb{R}^C$
stores historical predictions, as Eq.~\ref{eq:9}.
\begin{equation}
\label{eq:9}
\mathbf{P}_n \leftarrow m_{\mathrm{ELR}} \mathbf{P}_n + (1-m_{\mathrm{ELR}})\mathbf{p}_n,
\end{equation}
where $m_{\mathrm{ELR}}\in(0,1)$ is the ELR momentum.
The ELR loss is defined as Eq.~\ref{eq:10}.
\begin{equation}
\label{eq:10}
\mathcal{L}_{\mathrm{ELR}}
= -\lambda_{\mathrm{ELR}}
\mathbb{E}_{n}\Big[\log\big(1-\langle \mathbf{p}_n,\mathbf{P}_n\rangle\big)\Big],
\end{equation}
which penalizes deviations from early, stable predictions. This term is linearly warmed up during early epochs to avoid over-regularization.

\noindent\textbf{Fourier-Early Learning Regularization in Frequency Domain.}
While ELR in the time domain regularizes prediction probabilities based on time domain representations, it does not explicitly constrain the regularization from periodical patterns of biomedical sequence representations.
To further stabilize learning dynamics at the sequence level, we introduce
a early learning regularization in frequency domain, \ie, \textbf{FourierELR}.

Let $\mathbf{F} \in \mathbb{R}^{B \times T \times D}$ denote the sequence-level latent
representation extracted by the encoder $g(\cdot)$, where $B$ indicates batch size, $T$ indicates the temporal length, and $D$ is the feature
dimension.
We compute its frequency-domain magnitude as Eq.~\ref{eq:11}.
\begin{equation}
\label{eq:11}
\mathbf{M} = \lvert \mathrm{rFFT}(\mathbf{F}) \rvert,
\end{equation}
where $\mathrm{rFFT}$ computes the discrete Fourier transform for real input along the time dimension.
An EMA buffer $\mathbf{Q}_b$ is maintained for each sequence, where $b$ indicates the $b-$th sample in a batch, as Eq.~\ref{eq:12}.
\begin{equation}
\label{eq:12}
\mathbf{Q}_b \leftarrow m_f \mathbf{Q}_b + (1-m_f) \mathbf{M}_b,
\end{equation}
where $m_f$ is the \textbf{FourierELR} momentum.
The \textbf{FourierELR} loss is defined as Eq.~\ref{eq:13}.
\begin{equation}
\label{eq:13}
\mathcal{L}_{\mathrm{FELR}}
= \lambda_f
\mathbb{E}_{b}\Big[
1-\max\big(0,\cos(\tilde{\mathbf{M}}_b,\tilde{\mathbf{Q}}_b)\big)
\Big],
\end{equation}
where $\tilde{\mathbf{M}}_b$ and $\tilde{\mathbf{Q}}_b$ denote $\ell_2$-normalized, vectorized spectra, and $\lambda_f$ controls the strength of \textbf{FourierELR}.

The proposed regularizer is seamlessly incorporated into standard mini-batch training. Specifically, for each input sequence, the encoder produces a sequence-level latent representation, which is transformed into the frequency domain by applying $\mathrm{rFFT}$ along the temporal axis. The resulting magnitude spectrum is then used to update an EMA buffer that maintains historical frequency representations. During each training iteration, the current frequency representation is compared against its corresponding historical estimate via cosine similarity, and the resulting discrepancy is added to the overall training objective.

Notably, \textbf{FourierELR} introduces no additional optimization stages and does not modify the underlying training procedure. The regularization term is optimized jointly with all other loss components through standard backpropagation.

\noindent\textbf{Forward-Free Classification with Confidence Diversity Regularization (FF-CDR).}
A common approach to learning with class-dependent label noise is forward loss correction, which relies on estimating an explicit noise transition matrix to relate clean and noisy labels~\cite{patrini2017making}. However, accurately estimating this matrix is challenging in practice, and model training can be highly sensitive to estimation errors, which is very critical when we deal with domain shifts for biomedical data without reliable labels. To avoid these issues, we propose a forward-free classification objective that does not require an explicit transition matrix and instead regularizes model predictions directly.

In the context of sleep staging, the strong temporal dependencies and spectral redundancy inherent in multi-modal EEG and EOG signals, when coupled with noisy and imbalanced labels, can induce degenerate and biased predictions under standard cross-entropy training.

To tackle this problem, we introduce a novel Confidence Diversity Regularization (CDR). CDR promotes confident model predictions while simultaneously preserving output diversity, and is, to the best of our knowledge, the first approach empirically shown to effectively address label noise in biomedical data. To achieve CDR, InfoMax~\cite{hjelm2018learning} is adopted to maximize the mutual information.
By promoting stable and balanced predictions under noisy supervision, this objective provides an effective forward-free alternative for learning with noisy labels in sleep staging, especially benefiting the reliability of \textbf{JTF-ELR}.

Let $\mathbf{z} \in \mathbb{R}^{B\times T \times C}$ denote the logits,
where $B$ is the batch size, $T$ denotes the sequence length, and $C$ indicates the number of classes.
We reshape the logits into token-level representations and denote by
$\mathbf{z}_n \in \mathbb{R}^{C}$ the logits corresponding to token $n$, where $n=1,\ldots,B*T$.
The predicted probability vector is defined as
\begin{equation}
\mathbf{p}_n = \mathrm{SoftMax}(\mathbf{z}_n),
\end{equation}
where $p_{n,c}$ denotes the predicted probability that token $n$ belongs to class $c$.

The classification loss with CDR is reformulated as Eq.~\ref{eq:14}.
\begin{equation}
\label{eq:14}
\mathcal{L}_{\mathrm{FF\text{-}CDR}}
=
\mathcal{L}_{\mathrm{CE}}
+ \lambda_{\mathrm{IM}}
\Big(
\mathbb{E}_{n}\left[H(\mathbf{p}_b)\right] - H(\bar{\mathbf{p}})
\Big),
\end{equation}
where $b$ indexes a sequence-level sample in a mini-batch,
$H(\mathbf{p}_b) = -\sum_{c=0}^{C-1} \mathbf{p}_{(b,c)} \log \mathbf{p}_{(b,c)}$ is the prediction entropy,
$\bar{\mathbf{p}} = \frac{1}{B}\sum_{b=0}^{B-1} \mathbf{p}_b$ denotes the batch-mean prediction,
and $\lambda_{\mathrm{IM}}$ is a trade-off coefficient.

\noindent\textbf{Overall objective.}
The final training objective is as Eq.~\ref{eq:15}.
\begin{equation}
\label{eq:15}
\begin{aligned}
\mathcal{L}_{\text{total}}
=&\;\lambda_{\mathrm{ce}}\mathcal{L}_{\mathrm{FF\text{-}CDR}}
+ \alpha(t)\mathcal{L}_{\mathrm{ELR}}
+ \mathcal{L}_{\mathrm{FELR}} \\
&+ \lambda_1\mathcal{L}_{\mathrm{rec}}
+ \lambda_2\mathcal{L}_{\mathrm{ep}}
+ \lambda_3\mathcal{L}_{\mathrm{seq}},
\end{aligned}
\end{equation}
where $\lambda_{\mathrm{ce}}$ controls the forward-free classification term, $\alpha(t)\in[0,1]$ is an epoch-dependent warm-up coefficient that linearly increases the regularization weight during training, and $\lambda_1$, $\lambda_2$, and $\lambda_3$ are weighting coefficients.

\section{Experiments}
\subsection{Datasets}

We follow the preprocessing protocol of SleepDG~\cite{wang2024generalizable} and evaluate our approach and baseline methods on $5$ public datasets: \textbf{Sleep-EDFx}~\cite{Kemp2000AnalysisOA, Goldberger2000PhysioBankPA}, \textbf{HMC}~\cite{alvarez2021inter}, \textbf{ISRUC}~\cite{khalighi2016isruc}, \textbf{SHHS}~\cite{zhang2018national,quan1997sleep}, and \textbf{P2018}~\cite{ghassemi2018you}. One EEG channel and one EOG channel are used per dataset to ensure cross-dataset following~\cite{wang2024generalizable}. Sleep-EDFx and SHHS were originally labeled under R\&K~\cite{wolpert1969manual} and later aligned with AASM guidelines~\cite{berry2012aasm}. Signals were band-pass filtered (0.3--35Hz), resampled to 100Hz, Z-score normalized, and for Sleep-EDFx, trimmed to relevant PSG segments~\cite{Supratak2017DeepSleepNetAM}.

\subsection{Implementation Details}
For multi-source cross-dataset domain generalization, $4$ datasets served as source domains and $1$ dataset was used as target domain, which was excluded during training and validation, following~\cite{wang2022generalizing}. \textbf{FF-TRUST} was implemented in PyTorch using Adam with a learning rate of $1.2 \times 10^{-3}$, weight decay $1 \times 10^{-4}$, and $\lambda_1=\lambda_2=\lambda_3=0.5$. Models were trained for $50$ epochs with batch size $32$, dropout $0.1$, $L=20$, and $d=512$.
Hyperparameters were tuned on source-domain validation sets for different Noise Rates (NR). In the domain-invariant feature learning stage, $\lambda_{\mathrm{CE}}$ was set to $1.0$ for NR $\in \{0.2,0.4\}$ and $1.1$ for NR $=0.6$, while $\lambda_{\mathrm{IM}}$ increased from $0.02$ to $0.2$ as NR rose from $0.2$ to $0.6$. For ELR, $\lambda_{\mathrm{ELR}}$ was set to $0.12$ (NR $\in \{0.2,0.4\}$) and $0.1$ (NR $=0.6$), with momentum $0.92$ and $0.9$, respectively. A linear ramp-up $\alpha(t)$ was applied from epoch $20$ to $T_{\mathrm{end}} \in \{25,30,35\}$ for NR $\in \{0.2,0.4,0.6\}$. For \textbf{JTF-ELR}, $\lambda_f$ was set to $0.08$ (NR $\in \{0.2,0.4\}$) and $0.1$ (NR $=0.6$), with $m_f=0.9$. The model has $6.5$M parameters, identical to SleepDG~\cite{wang2024generalizable}. Performance was evaluated using ACC and MF1 on $8$ NVIDIA RTX 1080 Ti.
\subsection{Analysis of the Benchmark}
\begin{table}[t!]
\caption{Performance comparison of \textbf{FF-TRUST} with existing LNL methods under symmetric and asymmetric label noise. \textit{\textbf{NR}}: noise rate, \textbf{I}: Sleep-EDFx, \textbf{II}: HMC, \textbf{III}: ISRUC, \textbf{IV}: SHHS, \textbf{V}: P2018.}
\label{tab:comparison_sym_asym_FourierELR}
\centering
\footnotesize
\setlength{\tabcolsep}{4.5pt}
\renewcommand{\arraystretch}{1.05}
\resizebox{\columnwidth}{!}{\begin{tabular}{llc*{6}{cc}}
\toprule

\multicolumn{2}{c}{\makecell{\textbf{Source Domain}\\$\rightarrow$ \textbf{Target Domain}}} & \multicolumn{1}{c}{} &
\multicolumn{2}{c}{\makecell{\textbf{II,III,IV,V}\\$\rightarrow$ \textbf{I}}} &
\multicolumn{2}{c}{\makecell{\textbf{I,III,IV,V}\\$\rightarrow$ \textbf{II}}} &
\multicolumn{2}{c}{\makecell{\textbf{I,II,IV,V}\\$\rightarrow$ \textbf{III}}} &
\multicolumn{2}{c}{\makecell{\textbf{I,II,III,V}\\$\rightarrow$ \textbf{IV}}} &
\multicolumn{2}{c}{\makecell{\textbf{I,II,III,IV}\\$\rightarrow$ \textbf{V}}} &
\multicolumn{2}{c}{\textbf{Average}} \\

\cmidrule(lr){1-3}\cmidrule(lr){4-5}\cmidrule(lr){6-7}\cmidrule(lr){8-9}
\cmidrule(lr){10-11}\cmidrule(lr){12-13}\cmidrule(lr){14-15}

\textbf{Method} & \textbf{Setting} & \textbf{\textit{NR}} &
\textbf{ACC} & \textbf{MF1} &
\textbf{ACC} & \textbf{MF1} &
\textbf{ACC} & \textbf{MF1} &
\textbf{ACC} & \textbf{MF1} &
\textbf{ACC} & \textbf{MF1} &
\textbf{ACC} & \textbf{MF1} \\
\midrule

\multirow{7}{*}{SleepDG}
& BASE & 0
& 76.10 & 70.88 & 73.25 & 70.70 & 77.38 & 73.97 & 75.77 & 67.70 & 73.88 & 69.54 & 75.28 & 70.56 \\
\cmidrule(lr){2-15}
& \multirow{3}{*}{sym.}
& \cellcolor{softred!10}0.2 & \cellcolor{softred!10}75.99 & \cellcolor{softred!10}70.68 & \cellcolor{softred!10}71.70 & \cellcolor{softred!10}69.84 & \cellcolor{softred!10}76.36 & \cellcolor{softred!10}72.71 & \cellcolor{softred!10}75.22 & \cellcolor{softred!10}67.40 & \cellcolor{softred!10}73.40 & \cellcolor{softred!10}68.94 & \cellcolor{softred!10}74.53 & \cellcolor{softred!10}69.91 \\
&& \cellcolor{softred!20}0.4 & \cellcolor{softred!20}75.59 & \cellcolor{softred!20}69.73 & \cellcolor{softred!20}70.40 & \cellcolor{softred!20}68.77 & \cellcolor{softred!20}75.52 & \cellcolor{softred!20}71.79 & \cellcolor{softred!20}75.11 & \cellcolor{softred!20}66.96 & \cellcolor{softred!20}72.64 & \cellcolor{softred!20}68.18 & \cellcolor{softred!20}73.85 & \cellcolor{softred!20}69.09 \\
&& \cellcolor{softred!35}0.6 & \cellcolor{softred!35}70.43 & \cellcolor{softred!35}64.84 & \cellcolor{softred!35}65.87 & \cellcolor{softred!35}62.09 & \cellcolor{softred!35}73.03 & \cellcolor{softred!35}66.32 & \cellcolor{softred!35}73.04 & \cellcolor{softred!35}63.71 & \cellcolor{softred!35}70.84 & \cellcolor{softred!35}63.36 & \cellcolor{softred!35}70.64 & \cellcolor{softred!35}64.06 \\
\cmidrule(lr){2-15}
& \multirow{3}{*}{asym.}
& \cellcolor{paperblue!10}0.2 & \cellcolor{paperblue!10}76.14 & \cellcolor{paperblue!10}70.75 & \cellcolor{paperblue!10}73.28 & \cellcolor{paperblue!10}70.89 & \cellcolor{paperblue!10}77.02 & \cellcolor{paperblue!10}73.39 & \cellcolor{paperblue!10}74.40 & \cellcolor{paperblue!10}66.78 & \cellcolor{paperblue!10}73.33 & \cellcolor{paperblue!10}69.45 & \cellcolor{paperblue!10}74.83 & \cellcolor{paperblue!10}70.25 \\
&& \cellcolor{paperblue!20}0.4 & \cellcolor{paperblue!20}74.32 & \cellcolor{paperblue!20}68.75 & \cellcolor{paperblue!20}72.11 & \cellcolor{paperblue!20}70.22 & \cellcolor{paperblue!20}75.78 & \cellcolor{paperblue!20}71.89 & \cellcolor{paperblue!20}73.65 & \cellcolor{paperblue!20}66.32 & \cellcolor{paperblue!20}73.05 & \cellcolor{paperblue!20}69.30 & \cellcolor{paperblue!20}73.78 & \cellcolor{paperblue!20}69.30 \\
&& \cellcolor{paperblue!35}0.6 & \cellcolor{paperblue!35}67.28 & \cellcolor{paperblue!35}62.66 & \cellcolor{paperblue!35}57.85 & \cellcolor{paperblue!35}54.01 & \cellcolor{paperblue!35}63.34 & \cellcolor{paperblue!35}58.60 & \cellcolor{paperblue!35}68.90 & \cellcolor{paperblue!35}62.64 & \cellcolor{paperblue!35}66.88 & \cellcolor{paperblue!35}66.64 & \cellcolor{paperblue!35}64.85 & \cellcolor{paperblue!35}60.91 \\
\midrule

\multirow{6}{*}{NPN}
& \multirow{3}{*}{sym.}
& \cellcolor{softred!10}0.2 & \cellcolor{softred!10}73.41 & \cellcolor{softred!10}68.06 & \cellcolor{softred!10}68.83 & \cellcolor{softred!10}65.64 & \cellcolor{softred!10}75.05 & \cellcolor{softred!10}66.37 & \cellcolor{softred!10}74.34 & \cellcolor{softred!10}65.53 & \cellcolor{softred!10}72.94 & \cellcolor{softred!10}68.40 & \cellcolor{softred!10}72.91 & \cellcolor{softred!10}66.80 \\
& & \cellcolor{softred!20}0.4 & \cellcolor{softred!20}73.30 & \cellcolor{softred!20}66.43 & \cellcolor{softred!20}68.00 & \cellcolor{softred!20}65.49 & \cellcolor{softred!20}72.56 & \cellcolor{softred!20}67.66 & \cellcolor{softred!20}72.72 & \cellcolor{softred!20}63.72 & \cellcolor{softred!20}71.73 & \cellcolor{softred!20}67.27 & \cellcolor{softred!20}71.66 & \cellcolor{softred!20}66.11 \\
& & \cellcolor{softred!35}0.6 & \cellcolor{softred!35}66.25 & \cellcolor{softred!35}61.55 & \cellcolor{softred!35}64.21 & \cellcolor{softred!35}60.33 & \cellcolor{softred!35}64.79 & \cellcolor{softred!35}60.39 & \cellcolor{softred!35}70.61 & \cellcolor{softred!35}62.20 & \cellcolor{softred!35}69.53 & \cellcolor{softred!35}64.74 & \cellcolor{softred!35}67.08 & \cellcolor{softred!35}61.84 \\
\cmidrule(lr){2-15}
& \multirow{3}{*}{asym.}
& \cellcolor{paperblue!10}0.2 & \cellcolor{paperblue!10}74.86 & \cellcolor{paperblue!10}68.69 & \cellcolor{paperblue!10}72.61 & \cellcolor{paperblue!10}69.79 & \cellcolor{paperblue!10}75.61 & \cellcolor{paperblue!10}71.22 & \cellcolor{paperblue!10}73.97 & \cellcolor{paperblue!10}63.62 & \cellcolor{paperblue!10}72.03 & \cellcolor{paperblue!10}68.85 & \cellcolor{paperblue!10}73.82 & \cellcolor{paperblue!10}68.43 \\
& & \cellcolor{paperblue!20}0.4 & \cellcolor{paperblue!20}73.43 & \cellcolor{paperblue!20}67.45 & \cellcolor{paperblue!20}70.30 & \cellcolor{paperblue!20}67.85 & \cellcolor{paperblue!20}74.87 & \cellcolor{paperblue!20}70.58 & \cellcolor{paperblue!20}73.86 & \cellcolor{paperblue!20}64.22 & \cellcolor{paperblue!20}71.44 & \cellcolor{paperblue!20}67.81 & \cellcolor{paperblue!20}72.78 & \cellcolor{paperblue!20}67.58 \\
& & \cellcolor{paperblue!35}0.6 & \cellcolor{paperblue!35}57.89 & \cellcolor{paperblue!35}56.50 & \cellcolor{paperblue!35}59.00 & \cellcolor{paperblue!35}55.58 & \cellcolor{paperblue!35}62.97 & \cellcolor{paperblue!35}59.69 & \cellcolor{paperblue!35}67.82 & \cellcolor{paperblue!35}61.29 & \cellcolor{paperblue!35}67.29 & \cellcolor{paperblue!35}64.67 & \cellcolor{paperblue!35}62.99 & \cellcolor{paperblue!35}59.55 \\
\midrule

\multirow{6}{*}{JAL}
& \multirow{3}{*}{sym.}
& \cellcolor{softred!10}0.2 & \cellcolor{softred!10}76.63 & \cellcolor{softred!10}70.51 & \cellcolor{softred!10}\textbf{73.31} & \cellcolor{softred!10}70.07 & \cellcolor{softred!10}76.67 & \cellcolor{softred!10}72.41 & \cellcolor{softred!10}74.46 & \cellcolor{softred!10}66.16 & \cellcolor{softred!10}73.06 & \cellcolor{softred!10}68.64 & \cellcolor{softred!10}74.83 & \cellcolor{softred!10}69.56 \\
& & \cellcolor{softred!20}0.4 & \cellcolor{softred!20}73.57 & \cellcolor{softred!20}67.18 & \cellcolor{softred!20}70.37 & \cellcolor{softred!20}68.12 & \cellcolor{softred!20}75.27 & \cellcolor{softred!20}71.06 & \cellcolor{softred!20}73.78 & \cellcolor{softred!20}65.03 & \cellcolor{softred!20}71.59 & \cellcolor{softred!20}66.71 & \cellcolor{softred!20}72.92 & \cellcolor{softred!20}67.62 \\
& & \cellcolor{softred!35}0.6 & \cellcolor{softred!35}71.93 & \cellcolor{softred!35}65.47 & \cellcolor{softred!35}66.21 & \cellcolor{softred!35}63.64 & \cellcolor{softred!35}73.92 & \cellcolor{softred!35}69.37 & \cellcolor{softred!35}73.86 & \cellcolor{softred!35}65.78 & \cellcolor{softred!35}71.38 & \cellcolor{softred!35}66.15 & \cellcolor{softred!35}71.46 & \cellcolor{softred!35}66.08 \\
\cmidrule(lr){2-15}
& \multirow{3}{*}{asym.}
& \cellcolor{paperblue!10}0.2 & \cellcolor{paperblue!10}74.86 & \cellcolor{paperblue!10}68.69 & \cellcolor{paperblue!10}72.61 & \cellcolor{paperblue!10}69.79 & \cellcolor{paperblue!10}76.79 & \cellcolor{paperblue!10}73.10 & \cellcolor{paperblue!10}73.43 & \cellcolor{paperblue!10}64.93 & \cellcolor{paperblue!10}73.06 & \cellcolor{paperblue!10}68.64 & \cellcolor{paperblue!10}74.15 & \cellcolor{paperblue!10}69.03 \\
& & \cellcolor{paperblue!20}0.4 & \cellcolor{paperblue!20}74.59 & \cellcolor{paperblue!20}67.54 & \cellcolor{paperblue!20}71.28 & \cellcolor{paperblue!20}68.07 & \cellcolor{paperblue!20}75.39 & \cellcolor{paperblue!20}70.10 & \cellcolor{paperblue!20}72.74 & \cellcolor{paperblue!20}63.69 & \cellcolor{paperblue!20}71.66 & \cellcolor{paperblue!20}67.05 & \cellcolor{paperblue!20}73.13 & \cellcolor{paperblue!20}67.29 \\
& & \cellcolor{paperblue!35}0.6 & \cellcolor{paperblue!35}70.35 & \cellcolor{paperblue!35}60.65 & \cellcolor{paperblue!35}55.64 & \cellcolor{paperblue!35}54.66 & \cellcolor{paperblue!35}67.05 & \cellcolor{paperblue!35}62.74 & \cellcolor{paperblue!35}64.15 & \cellcolor{paperblue!35}57.35 & \cellcolor{paperblue!35}66.44 & \cellcolor{paperblue!35}62.35 & \cellcolor{paperblue!35}64.73 & \cellcolor{paperblue!35}59.55 \\
\midrule

\multirow{6}{*}{DISC}
& \multirow{3}{*}{sym.}
& \cellcolor{softred!10}0.2 & \cellcolor{softred!10}76.75 & \cellcolor{softred!10}71.33 & \cellcolor{softred!10}72.42 & \cellcolor{softred!10}70.00 & \cellcolor{softred!10}76.99 & \cellcolor{softred!10}73.32 & \cellcolor{softred!10}74.73 & \cellcolor{softred!10}67.01 & \cellcolor{softred!10}73.77 & \cellcolor{softred!10}69.47 & \cellcolor{softred!10}74.93 & \cellcolor{softred!10}70.23 \\
& & \cellcolor{softred!20}0.4 & \cellcolor{softred!20}75.99 & \cellcolor{softred!20}70.02 & \cellcolor{softred!20}71.19 & \cellcolor{softred!20}69.04 & \cellcolor{softred!20}75.54 & \cellcolor{softred!20}71.24 & \cellcolor{softred!20}74.61 & \cellcolor{softred!20}66.32 & \cellcolor{softred!20}73.18 & \cellcolor{softred!20}69.37 & \cellcolor{softred!20}74.10 & \cellcolor{softred!20}69.20 \\
& & \cellcolor{softred!35}0.6 & \cellcolor{softred!35}71.78 & \cellcolor{softred!35}64.82 & \cellcolor{softred!35}68.17 & \cellcolor{softred!35}65.24 & \cellcolor{softred!35}72.95 & \cellcolor{softred!35}67.12 & \cellcolor{softred!35}73.77 & \cellcolor{softred!35}64.14 & \cellcolor{softred!35}71.77 & \cellcolor{softred!35}66.04 & \cellcolor{softred!35}71.69 & \cellcolor{softred!35}65.47 \\
\cmidrule(lr){2-15}
& \multirow{3}{*}{asym.}
& \cellcolor{paperblue!10}0.2 & \cellcolor{paperblue!10}76.44 & \cellcolor{paperblue!10}70.81 & \cellcolor{paperblue!10}72.28 & \cellcolor{paperblue!10}70.27 & \cellcolor{paperblue!10}76.82 & \cellcolor{paperblue!10}73.07 & \cellcolor{paperblue!10}73.53 & \cellcolor{paperblue!10}65.91 & \cellcolor{paperblue!10}74.50 & \cellcolor{paperblue!10}70.39 & \cellcolor{paperblue!10}74.71 & \cellcolor{paperblue!10}70.09 \\
& & \cellcolor{paperblue!20}0.4 & \cellcolor{paperblue!20}75.16 & \cellcolor{paperblue!20}69.75 & \cellcolor{paperblue!20}70.93 & \cellcolor{paperblue!20}68.82 & \cellcolor{paperblue!20}75.71 & \cellcolor{paperblue!20}72.18 & \cellcolor{paperblue!20}73.18 & \cellcolor{paperblue!20}65.90 & \cellcolor{paperblue!20}73.30 & \cellcolor{paperblue!20}69.84 & \cellcolor{paperblue!20}73.66 & \cellcolor{paperblue!20}69.30 \\
& & \cellcolor{paperblue!35}0.6 & \cellcolor{paperblue!35}68.56 & \cellcolor{paperblue!35}62.76 & \cellcolor{paperblue!35}59.37 & \cellcolor{paperblue!35}56.04 & \cellcolor{paperblue!35}62.47 & \cellcolor{paperblue!35}56.56 & \cellcolor{paperblue!35}66.49 & \cellcolor{paperblue!35}60.00 & \cellcolor{paperblue!35}65.84 & \cellcolor{paperblue!35}62.58 & \cellcolor{paperblue!35}64.55 & \cellcolor{paperblue!35}59.59 \\
\midrule

\multirow{6}{*}{FedDSHAR}
& \multirow{3}{*}{sym.}
& \cellcolor{softred!10}0.2 & \cellcolor{softred!10}76.86 & \cellcolor{softred!10}71.00 & \cellcolor{softred!10}70.20 & \cellcolor{softred!10}67.96 & \cellcolor{softred!10}76.95 & \cellcolor{softred!10}72.38 & \cellcolor{softred!10}74.32 & \cellcolor{softred!10}66.21 & \cellcolor{softred!10}73.00 & \cellcolor{softred!10}68.15 & \cellcolor{softred!10}74.27 & \cellcolor{softred!10}69.14 \\
& & \cellcolor{softred!20}0.4 & \cellcolor{softred!20}75.97 & \cellcolor{softred!20}69.82 & \cellcolor{softred!20}70.36 & \cellcolor{softred!20}68.39 & \cellcolor{softred!20}76.63 & \cellcolor{softred!20}72.77 & \cellcolor{softred!20}73.48 & \cellcolor{softred!20}64.40 & \cellcolor{softred!20}72.87 & \cellcolor{softred!20}68.61 & \cellcolor{softred!20}73.86 & \cellcolor{softred!20}68.80 \\
& & \cellcolor{softred!35}0.6 & \cellcolor{softred!35}72.46 & \cellcolor{softred!35}65.98 & \cellcolor{softred!35}67.22 & \cellcolor{softred!35}64.22 & \cellcolor{softred!35}73.61 & \cellcolor{softred!35}68.51 & \cellcolor{softred!35}73.89 & \cellcolor{softred!35}65.47 & \cellcolor{softred!35}70.95 & \cellcolor{softred!35}64.96 & \cellcolor{softred!35}71.63 & \cellcolor{softred!35}65.83 \\
\cmidrule(lr){2-15}
& \multirow{3}{*}{asym.}
& \cellcolor{paperblue!10}0.2 & \cellcolor{paperblue!10}76.96 & \cellcolor{paperblue!10}70.84 & \cellcolor{paperblue!10}71.26 & \cellcolor{paperblue!10}68.90 & \cellcolor{paperblue!10}76.80 & \cellcolor{paperblue!10}72.74 & \cellcolor{paperblue!10}75.18 & \cellcolor{paperblue!10}67.13 & \cellcolor{paperblue!10}72.87 & \cellcolor{paperblue!10}68.09 & \cellcolor{paperblue!10}74.61 & \cellcolor{paperblue!10}69.54 \\
& & \cellcolor{paperblue!20}0.4 & \cellcolor{paperblue!20}76.08 & \cellcolor{paperblue!20}70.11 & \cellcolor{paperblue!20}70.81 & \cellcolor{paperblue!20}68.87 & \cellcolor{paperblue!20}75.50 & \cellcolor{paperblue!20}71.21 & \cellcolor{paperblue!20}73.01 & \cellcolor{paperblue!20}65.14 & \cellcolor{paperblue!20}73.86 & \cellcolor{paperblue!20}68.70 & \cellcolor{paperblue!20}73.85 & \cellcolor{paperblue!20}68.81 \\
& & \cellcolor{paperblue!35}0.6 & \cellcolor{paperblue!35}70.32 & \cellcolor{paperblue!35}64.35 & \cellcolor{paperblue!35}60.59 & \cellcolor{paperblue!35}58.01 & \cellcolor{paperblue!35}69.93 & \cellcolor{paperblue!35}64.83 & \cellcolor{paperblue!35}67.37 & \cellcolor{paperblue!35}60.01 & \cellcolor{paperblue!35}68.66 & \cellcolor{paperblue!35}64.90 & \cellcolor{paperblue!35}67.37 & \cellcolor{paperblue!35}62.42 \\
\midrule

\multirow{6}{*}{ELR}
& \multirow{3}{*}{sym.}
& \cellcolor{softred!10}0.2 & \cellcolor{softred!10}76.95 & \cellcolor{softred!10}71.56 & \cellcolor{softred!10}72.01 & \cellcolor{softred!10}70.11 & \cellcolor{softred!10}76.92 & \cellcolor{softred!10}73.41 & \cellcolor{softred!10}74.45 & \cellcolor{softred!10}66.32 & \cellcolor{softred!10}73.61 & \cellcolor{softred!10}69.08 & \cellcolor{softred!10}74.83 & \cellcolor{softred!10}70.10 \\
& & \cellcolor{softred!20}0.4 & \cellcolor{softred!20}76.70 & \cellcolor{softred!20}71.21 & \cellcolor{softred!20}71.31 & \cellcolor{softred!20}69.03 & \cellcolor{softred!20}75.39 & \cellcolor{softred!20}72.04 & \cellcolor{softred!20}74.91 & \cellcolor{softred!20}66.36 & \cellcolor{softred!20}72.58 & \cellcolor{softred!20}68.12 & \cellcolor{softred!20}74.18 & \cellcolor{softred!20}69.35 \\
& & \cellcolor{softred!35}0.6 & \cellcolor{softred!35}71.54 & \cellcolor{softred!35}65.20 & \cellcolor{softred!35}68.37 & \cellcolor{softred!35}66.15 & \cellcolor{softred!35}73.21 & \cellcolor{softred!35}68.73 & \cellcolor{softred!35}72.01 & \cellcolor{softred!35}65.80 & \cellcolor{softred!35}70.82 & \cellcolor{softred!35}65.63 & \cellcolor{softred!35}71.19 & \cellcolor{softred!35}66.30 \\
\cmidrule(lr){2-15}
& \multirow{3}{*}{asym.}
& \cellcolor{paperblue!10}0.2 & \cellcolor{paperblue!10}75.83 & \cellcolor{paperblue!10}70.18 & \cellcolor{paperblue!10}73.00 & \cellcolor{paperblue!10}71.18 & \cellcolor{paperblue!10}75.96 & \cellcolor{paperblue!10}72.16 & \cellcolor{paperblue!10}74.61 & \cellcolor{paperblue!10}66.62 & \cellcolor{paperblue!10}73.05 & \cellcolor{paperblue!10}69.61 & \cellcolor{paperblue!10}74.49 & \cellcolor{paperblue!10}69.95 \\
& & \cellcolor{paperblue!20}0.4 & \cellcolor{paperblue!20}75.27 & \cellcolor{paperblue!20}69.98 & \cellcolor{paperblue!20}\textbf{72.98} & \cellcolor{paperblue!20}\textbf{71.09} & \cellcolor{paperblue!20}76.35 & \cellcolor{paperblue!20}72.40 & \cellcolor{paperblue!20}72.99 & \cellcolor{paperblue!20}65.58 & \cellcolor{paperblue!20}\textbf{74.13} & \cellcolor{paperblue!20}70.43 & \cellcolor{paperblue!20}74.34 & \cellcolor{paperblue!20}69.90 \\
& & \cellcolor{paperblue!35}0.6 & \cellcolor{paperblue!35}67.97 & \cellcolor{paperblue!35}63.00 & \cellcolor{paperblue!35}62.39 & \cellcolor{paperblue!35}60.21 & \cellcolor{paperblue!35}68.66 & \cellcolor{paperblue!35}64.76 & \cellcolor{paperblue!35}64.98 & \cellcolor{paperblue!35}59.87 & \cellcolor{paperblue!35}68.14 & \cellcolor{paperblue!35}66.15 & \cellcolor{paperblue!35}66.43 & \cellcolor{paperblue!35}62.80 \\
\midrule

\multirow{6}{*}{\textbf{FF-TRUST (ours)}}
& \multirow{3}{*}{sym.}
& \cellcolor{softred!10}0.2 & \cellcolor{softred!10}\textbf{77.18} & \cellcolor{softred!10}\textbf{71.70} & \cellcolor{softred!10}72.44 & \cellcolor{softred!10}\textbf{70.79} & \cellcolor{softred!10}\textbf{78.02} & \cellcolor{softred!10}\textbf{75.11} & \cellcolor{softred!10}\textbf{76.03} & \cellcolor{softred!10}\textbf{67.88} & \cellcolor{softred!10}\textbf{74.08} & \cellcolor{softred!10}\textbf{69.92} & \cellcolor{softred!10}\textbf{75.55} & \cellcolor{softred!10}\textbf{71.08} \\
& & \cellcolor{softred!20}0.4 & \cellcolor{softred!20}\textbf{76.71} & \cellcolor{softred!20}\textbf{71.58} & \cellcolor{softred!20}\textbf{71.90} & \cellcolor{softred!20}\textbf{70.24} & \cellcolor{softred!20}\textbf{77.05} & \cellcolor{softred!20}\textbf{73.79} & \cellcolor{softred!20}\textbf{75.43} & \cellcolor{softred!20}\textbf{67.38} & \cellcolor{softred!20}\textbf{73.26} & \cellcolor{softred!20}\textbf{69.68} & \cellcolor{softred!20}\textbf{74.87} & \cellcolor{softred!20}\textbf{70.53} \\
& & \cellcolor{softred!35}0.6 & \cellcolor{softred!35}\textbf{75.13} & \cellcolor{softred!35}\textbf{68.95} & \cellcolor{softred!35}\textbf{70.80} & \cellcolor{softred!35}\textbf{68.79} & \cellcolor{softred!35}\textbf{74.63} & \cellcolor{softred!35}\textbf{71.02} & \cellcolor{softred!35}\textbf{74.99} & \cellcolor{softred!35}\textbf{66.20} & \cellcolor{softred!35}\textbf{72.51} & \cellcolor{softred!35}\textbf{67.40} & \cellcolor{softred!35}\textbf{73.61} & \cellcolor{softred!35}\textbf{68.47} \\
\cmidrule(lr){2-15}
& \multirow{3}{*}{asym.}
& \cellcolor{paperblue!10}0.2 & \cellcolor{paperblue!10}\textbf{77.30} & \cellcolor{paperblue!10}\textbf{72.01} & \cellcolor{paperblue!10}\textbf{73.50} & \cellcolor{paperblue!10}\textbf{71.16} & \cellcolor{paperblue!10}\textbf{77.65} & \cellcolor{paperblue!10}\textbf{74.22} & \cellcolor{paperblue!10}\textbf{75.84} & \cellcolor{paperblue!10}\textbf{67.87} & \cellcolor{paperblue!10}\textbf{74.66} & \cellcolor{paperblue!10}\textbf{71.11} & \cellcolor{paperblue!10}\textbf{75.79} & \cellcolor{paperblue!10}\textbf{71.27} \\
& & \cellcolor{paperblue!20}0.4 & \cellcolor{paperblue!20}\textbf{76.79} & \cellcolor{paperblue!20}\textbf{70.96} & \cellcolor{paperblue!20}72.71 & \cellcolor{paperblue!20}70.52 & \cellcolor{paperblue!20}\textbf{76.57} & \cellcolor{paperblue!20}\textbf{73.00} & \cellcolor{paperblue!20}\textbf{75.39} & \cellcolor{paperblue!20}\textbf{67.42} & \cellcolor{paperblue!20}73.94 & \cellcolor{paperblue!20}\textbf{70.51} & \cellcolor{paperblue!20}\textbf{75.08} & \cellcolor{paperblue!20}\textbf{70.48} \\
& & \cellcolor{paperblue!35}0.6 & \cellcolor{paperblue!35}\textbf{71.88} & \cellcolor{paperblue!35}\textbf{66.55} & \cellcolor{paperblue!35}\textbf{69.59} & \cellcolor{paperblue!35}\textbf{64.79} & \cellcolor{paperblue!35}\textbf{72.20} & \cellcolor{paperblue!35}\textbf{67.83} & \cellcolor{paperblue!35}\textbf{74.34} & \cellcolor{paperblue!35}\textbf{65.48} & \cellcolor{paperblue!35}\textbf{70.94} & \cellcolor{paperblue!35}\textbf{69.09} & \cellcolor{paperblue!35}\textbf{71.79} & \cellcolor{paperblue!35}\textbf{66.75} \\
\bottomrule
\end{tabular}}
\end{table}

We evaluate \textbf{FF-TRUST} and $5$ LNL baselines built on the SleepDG backbone~\cite{wang2024generalizable} across $5$ sleep staging datasets under symmetric and asymmetric label noise (Table~\ref{tab:comparison_sym_asym_FourierELR}). As the noise rate increases from $0$ to $0.6$, both average accuracy and MF1 consistently decrease for all methods. While performance is comparable under both noise types at low and moderate levels, asymmetric label noise causes markedly larger degradation at high noise rates. Notably, our study addresses noise from the \emph{label perspective} rather than the data perspective.

\noindent\textbf{High-Noise Setting (NR = 0.6).}
\textbf{FF-TRUST} shows strong robustness under severe label noise, with only small gaps between symmetric and asymmetric noise at a rate of $0.6$ ($1.82\%$ in accuracy and $1.72\%$ in MF1). Moreover, \textbf{FF-TRUST} achieves the best overall performance among all methods, reaching $73.61\%$ ACC / $68.47\%$ MF1 under symmetric noise and $71.79\%$ ACC / $66.75\%$ MF1 under asymmetric noise.
Under asymmetric noise, \textbf{FF-TRUST} surpasses the strongest competing baseline, \textit{FedDSHAR}~\cite{lin2025feddshar}, by $4.42\%$ in ACC and $4.33\%$ in MF1, and further improves over SleepDG~\cite{wang2024generalizable} by $6.94\%$ and $5.84\%$, respectively.
In contrast, \textit{DISC}~\cite{li2023disc} exhibits the largest sensitivity to asymmetry noise, with ACC and MF1 gaps reaching $7.14\%$ and $5.88\%$, respectively. \textit{JAL}~\cite{wang2025joint} shows only marginal improvement over \textit{DISC}~\cite{li2023disc}, yet still suffers from considerable gaps of $6.73\%$ in ACC and $6.53\%$ in MF1.
\textit{NPN}~\cite{sheng2024adaptive} performs the worst among all methods, achieving only $67.08\%$ ACC and $61.84\%$ MF1 under symmetric noise, and further dropping to $62.99\%$ ACC and $59.55\%$ MF1 under asymmetric noise, as \textit{NPN}~\cite{sheng2024adaptive} is primarily designed for independent-sample settings, whereas sleep staging involves strong temporal dependencies, causing pseudo-label errors to propagate across consecutive epochs.

\noindent\textbf{Moderate-Noise Setting (NR = 0.4).}
We select $NR=0.4$ to serve as the moderate label noise situation. Comparing with \textit{SleepDG w/o label noise (BASE)}~\cite{wang2024generalizable}, symmetric and asymmetric label noise at $0.4$ result in $1.02\%$, and $1.30\%$ ACC degradations and $1.44\%$, and $1.18\%$ MF1 degradations.
Among all the leveraged baselines, \textit{ELR}~\cite{liu2020early} delivers the best performances with $74.18\%$, $74.34\%$ ACC and $69.35\%$, $69.90\%$ MF1. Most of the other baselines, \eg, \textit{NPN}~\cite{sheng2024adaptive}, \textit{JAL}~\cite{wang2025joint}, and \textit{DISC}~\cite{li2023disc}, deliver $<74.1\%$ ACC and $69.30\%$ MF1.
\textbf{FF-TRUST} continues to demonstrate strong and stable performance ($74.87\%$ and $75.08\%$ ACC, $70.53\%$ and $70.48\%$ MF1, for symmetric and asymmetric label noise, respectively), achieving the best overall average accuracy and MF1 under symmetric and asymmetric noise.

\noindent\textbf{Low-Noise Setting (NR = 0.2).}
When the noise rate is reduced to $0.2$, \textbf{FF-TRUST} achieves the best performance across all baselines. Notably, its average accuracy and MF1 under both symmetric ($75.55\%$ ACC, $71.08\%$ MF1) and asymmetric noise ($75.79\%$ ACC, $71.27\%$ MF1), and even exceed those of \textit{SleepDG w/o label noise (BASE)}~\cite{wang2024generalizable}, indicating that \textbf{FF-TRUST} can effectively mitigate mild label noise while preserving strong generalization performance. We further implement \textbf{FF-Trust} on one conventional sleep staging method, \ie, DeepSleepNet~\cite{Supratak2017DeepSleepNetAM}, to demonstrate its cross-backbone generalizability in Table~\ref{tab:deepsleep_sym_asym}.
In summary, \textbf{FF-TRUST} not only achieves the highest absolute performance under both symmetric and asymmetric label noise, but also exhibits remarkable insensitivity to noise type, especially in high-noise regimes, highlighting its strong robustness to structured class-dependent label corruption.

In our \textbf{FF-Trust}, \textbf{JTF-ELR} exploits the time-frequency early-learning dynamics of sequential data by constraining both time- and frequency-domain representations to remain consistent with their EMA-based historical estimates, thereby preventing memorization of noisy labels and stabilizing optimization.
In addition, \textbf{FF-CDR} avoids explicit noise transition modeling and instead promotes confident yet diverse predictions, reducing class-collapse and bias under noisy supervision, which jointly explains the superior robustness of \textbf{FF-TRUST} over existing baselines.
\begin{table}[t]
\caption{Performance comparison of \textbf{FF-TRUST} with existing model under symmetric and asymmetric label noise. \textit{\textbf{NR}}: noise rate, \textbf{I}: Sleep-EDFx, \textbf{II}: HMC, \textbf{III}: ISRUC, \textbf{IV}: SHHS, \textbf{V}: P2018.}
\centering
\footnotesize
\setlength{\tabcolsep}{4.5pt}
\renewcommand{\arraystretch}{1.05}
\resizebox{\columnwidth}{!}{\begin{tabular}{llc*{6}{cc}}
\toprule

\multicolumn{2}{c}{\makecell{\textbf{Source Domain}\\$\rightarrow$ \textbf{Target Domain}}} & \multicolumn{1}{c}{} &
\multicolumn{2}{c}{\makecell{\textbf{II,III,IV,V}\\$\rightarrow$ \textbf{I}}} &
\multicolumn{2}{c}{\makecell{\textbf{I,III,IV,V}\\$\rightarrow$ \textbf{II}}} &
\multicolumn{2}{c}{\makecell{\textbf{I,II,IV,V}\\$\rightarrow$ \textbf{III}}} &
\multicolumn{2}{c}{\makecell{\textbf{I,II,III,V}\\$\rightarrow$ \textbf{IV}}} &
\multicolumn{2}{c}{\makecell{\textbf{I,II,III,IV}\\$\rightarrow$ \textbf{V}}} &
\multicolumn{2}{c}{\textbf{Average}} \\

\cmidrule(lr){1-3}\cmidrule(lr){4-5}\cmidrule(lr){6-7}\cmidrule(lr){8-9}
\cmidrule(lr){10-11}\cmidrule(lr){12-13}\cmidrule(lr){14-15}

\textbf{Method} & \textbf{Setting} & \textbf{\textit{NR}} &
\textbf{ACC} & \textbf{MF1} &
\textbf{ACC} & \textbf{MF1} &
\textbf{ACC} & \textbf{MF1} &
\textbf{ACC} & \textbf{MF1} &
\textbf{ACC} & \textbf{MF1} &
\textbf{ACC} & \textbf{MF1} \\
\midrule

\multirow{7}{*}{DeepSleepNet}
& BASE & 0 & 74.10 & 68.69 & 70.28 & 67.75 & 77.47 & 73.65 & 74.62 & 67.22 & 71.23 & 66.66 & 73.54 & 68.79 \\
\cmidrule(lr){2-15}
& \multirow{3}{*}{sym.}
& \cellcolor{softred!10}0.2 & \cellcolor{softred!10}72.68 & \cellcolor{softred!10}67.79 & \cellcolor{softred!10}69.77 & \cellcolor{softred!10}67.37 & \cellcolor{softred!10}77.01 & \cellcolor{softred!10}72.80 & \cellcolor{softred!10}73.49 & \cellcolor{softred!10}65.89 & \cellcolor{softred!10}70.80 & \cellcolor{softred!10}66.75 & \cellcolor{softred!10}72.75 & \cellcolor{softred!10}68.12\\
&& \cellcolor{softred!20}0.4 & \cellcolor{softred!20}70.69 & \cellcolor{softred!20}65.44 & \cellcolor{softred!20}67.16 & \cellcolor{softred!20}64.85 & \cellcolor{softred!20}76.32 & \cellcolor{softred!20}71.28 & \cellcolor{softred!20}74.19 & \cellcolor{softred!20}66.33 & \cellcolor{softred!20}70.74 & \cellcolor{softred!20}65.39 & \cellcolor{softred!20}71.82 & \cellcolor{softred!20}66.66 \\
&& \cellcolor{softred!35}0.6 & \cellcolor{softred!35}69.60 & \cellcolor{softred!35}62.37 & \cellcolor{softred!35}65.37 & \cellcolor{softred!35}61.71 & \cellcolor{softred!35}74.40 & \cellcolor{softred!35}67.94 & \cellcolor{softred!35}73.27 & \cellcolor{softred!35}64.68 & \cellcolor{softred!35}69.23 & \cellcolor{softred!35}61.71 & \cellcolor{softred!35}70.37 & \cellcolor{softred!35}63.68 \\
\cmidrule(lr){2-15}
& \multirow{3}{*}{asym.}
& \cellcolor{paperblue!10}0.2 & \cellcolor{paperblue!10}73.07 & \cellcolor{paperblue!10}67.14 & \cellcolor{paperblue!10}68.69 & \cellcolor{paperblue!10}66.58 & \cellcolor{paperblue!10}76.96 & \cellcolor{paperblue!10}72.49 & \cellcolor{paperblue!10}73.78 & \cellcolor{paperblue!10}66.16 & \cellcolor{paperblue!10}70.90 & \cellcolor{paperblue!10}66.46 & \cellcolor{paperblue!10}72.68 & \cellcolor{paperblue!10}67.77 \\
&& \cellcolor{paperblue!20}0.4 & \cellcolor{paperblue!20}71.97 & \cellcolor{paperblue!20}67.29 & \cellcolor{paperblue!20}68.35 & \cellcolor{paperblue!20}66.58 & \cellcolor{paperblue!20}76.83 & \cellcolor{paperblue!20}72.54 & \cellcolor{paperblue!20}68.93 & \cellcolor{paperblue!20}61.41 & \cellcolor{paperblue!20}71.79 & \cellcolor{paperblue!20}68.10 & \cellcolor{paperblue!20}71.57 & \cellcolor{paperblue!20}67.18 \\
&& \cellcolor{paperblue!35}0.6 & \cellcolor{paperblue!35}57.90 & \cellcolor{paperblue!35}53.31 & \cellcolor{paperblue!35}58.32 & \cellcolor{paperblue!35}55.71 & \cellcolor{paperblue!35}66.93 & \cellcolor{paperblue!35}62.85 & \cellcolor{paperblue!35}58.56 & \cellcolor{paperblue!35}54.28 & \cellcolor{paperblue!35}64.54 & \cellcolor{paperblue!35}62.41 & \cellcolor{paperblue!35}61.25 & \cellcolor{paperblue!35}57.71 \\
\midrule

\multirow{6}{*}{\textbf{FF-TRUST (ours)}}
& \multirow{3}{*}{sym.}
& \cellcolor{softred!10}0.2 & \cellcolor{softred!10}\textbf{73.82} & \cellcolor{softred!10}\textbf{68.56} & \cellcolor{softred!10}70.25 & \cellcolor{softred!10}\textbf{68.52} & \cellcolor{softred!10}\textbf{77.43} & \cellcolor{softred!10}\textbf{73.87} & \cellcolor{softred!10}\textbf{74.05} & \cellcolor{softred!10}\textbf{66.51} & \cellcolor{softred!10}\textbf{71.41} & \cellcolor{softred!10}\textbf{67.61} & \cellcolor{softred!10}\textbf{73.39} & \cellcolor{softred!10}\textbf{69.01} \\
& & \cellcolor{softred!20}0.4 & \cellcolor{softred!20}\textbf{72.20} & \cellcolor{softred!20}\textbf{66.37} & \cellcolor{softred!20}\textbf{67.41} & \cellcolor{softred!20}\textbf{65.11} & \cellcolor{softred!20}\textbf{77.10} & \cellcolor{softred!20}\textbf{72.31} & \cellcolor{softred!20}\textbf{74.48} & \cellcolor{softred!20}\textbf{66.33} & \cellcolor{softred!20}\textbf{70.90} & \cellcolor{softred!20}\textbf{65.85} & \cellcolor{softred!20}\textbf{72.42} & \cellcolor{softred!20}\textbf{67.19} \\
& & \cellcolor{softred!35}0.6 & \cellcolor{softred!35}\textbf{70.62} & \cellcolor{softred!35}\textbf{64.24} & \cellcolor{softred!35}\textbf{66.96} & \cellcolor{softred!35}\textbf{64.35} & \cellcolor{softred!35}\textbf{75.82} & \cellcolor{softred!35}\textbf{70.48} & \cellcolor{softred!35}\textbf{73.73} & \cellcolor{softred!35}\textbf{65.14} & \cellcolor{softred!35}\textbf{69.77} & \cellcolor{softred!35}\textbf{64.90} & \cellcolor{softred!35}\textbf{71.38} & \cellcolor{softred!35}\textbf{65.82} \\
\cmidrule(lr){2-15}
& \multirow{3}{*}{asym.}
& \cellcolor{paperblue!10}0.2 & \cellcolor{paperblue!10}\textbf{73.08} & \cellcolor{paperblue!10}\textbf{68.07} & \cellcolor{paperblue!10}\textbf{69.85} & \cellcolor{paperblue!10}\textbf{67.80} & \cellcolor{paperblue!10}\textbf{76.87} & \cellcolor{paperblue!10}\textbf{72.74} & \cellcolor{paperblue!10}\textbf{73.95} & \cellcolor{paperblue!10}\textbf{66.35} & \cellcolor{paperblue!10}\textbf{72.14} & \cellcolor{paperblue!10}\textbf{67.20} & \cellcolor{paperblue!10}\textbf{73.18} & \cellcolor{paperblue!10}\textbf{68.43} \\
& & \cellcolor{paperblue!20}0.4 & \cellcolor{paperblue!20}\textbf{72.39} & \cellcolor{paperblue!20}\textbf{67.00} & \cellcolor{paperblue!20}\textbf{70.89} & \cellcolor{paperblue!20}\textbf{68.35} & \cellcolor{paperblue!20}\textbf{77.33} & \cellcolor{paperblue!20}\textbf{73.04} & \cellcolor{paperblue!20}\textbf{72.81} & \cellcolor{paperblue!20}\textbf{64.92} & \cellcolor{paperblue!20}\textbf{72.36} & \cellcolor{paperblue!20}\textbf{68.99} & \cellcolor{paperblue!20}\textbf{73.16} & \cellcolor{paperblue!20}\textbf{68.46} \\
& & \cellcolor{paperblue!35}0.6 & \cellcolor{paperblue!35}\textbf{68.32} & \cellcolor{paperblue!35}\textbf{62.81} & \cellcolor{paperblue!35}\textbf{61.15} & \cellcolor{paperblue!35}\textbf{60.10} & \cellcolor{paperblue!35}\textbf{72.83} & \cellcolor{paperblue!35}\textbf{69.67} & \cellcolor{paperblue!35}\textbf{67.93} & \cellcolor{paperblue!35}\textbf{60.86} & \cellcolor{paperblue!35}\textbf{66.68} & \cellcolor{paperblue!35}\textbf{65.49} & \cellcolor{paperblue!35}\textbf{67.38} & \cellcolor{paperblue!35}\textbf{63.79} \\
\bottomrule
\end{tabular}}
\label{tab:deepsleep_sym_asym}
\end{table}

\subsection{Analysis of Module Ablation}
\begin{table}[t]
\caption{Module ablation study under $0.6$ symmetric label noise. We verify the effectiveness of TimeELR, FourierELR and FF-CDR.}
\vskip-1ex
\label{tab:ablation}
\centering
\footnotesize
\setlength{\tabcolsep}{4pt}
\renewcommand{\arraystretch}{1.05}
\resizebox{\columnwidth}{!}{\begin{tabular}{ccc *{6}{cc}}
\toprule

\multicolumn{3}{c}{\textbf{Modules}} &
\multicolumn{2}{c}{\makecell{\textbf{II,III,IV,V}\\$\rightarrow$ \textbf{I}}} &
\multicolumn{2}{c}{\makecell{\textbf{I,III,IV,V}\\$\rightarrow$ \textbf{II}}} &
\multicolumn{2}{c}{\makecell{\textbf{I,II,IV,V}\\$\rightarrow$ \textbf{III}}} &
\multicolumn{2}{c}{\makecell{\textbf{I,II,III,V}\\$\rightarrow$ \textbf{IV}}} &
\multicolumn{2}{c}{\makecell{\textbf{I,II,III,IV}\\$\rightarrow$ \textbf{V}}} &
\multicolumn{2}{c}{\textbf{Average}} \\

\cmidrule(lr){1-3}
\cmidrule(lr){4-5}\cmidrule(lr){6-7}\cmidrule(lr){8-9}
\cmidrule(lr){10-11}\cmidrule(lr){12-13}\cmidrule(lr){14-15}

\textbf{TimeELR} & \textbf{FourierELR} & \textbf{FF-CDR} &
\textbf{ACC} & \textbf{MF1} &
\textbf{ACC} & \textbf{MF1} &
\textbf{ACC} & \textbf{MF1} &
\textbf{ACC} & \textbf{MF1} &
\textbf{ACC} & \textbf{MF1} &
\textbf{ACC} & \textbf{MF1} \\

\midrule

 &  &
& 70.43 & 64.84 & 65.87 & 62.09 & 73.03 & 66.32 & 73.04 & 63.71 & 70.84 & 63.36 & 70.64 & 64.06 \\

\checkmark &  &
& 71.54 & 65.20 & 68.37 & 66.15 & 73.21 & 68.73 & 72.01 & 65.80 & 70.82 & 65.63 & 71.19 & 66.30 \\

 &  & \checkmark
& 71.44 & 65.83 & 68.41 & 66.95 & 73.37 & 69.50 & 74.18 & 65.74 & 70.62 & 66.25 & 71.60 & 66.85 \\

 & \checkmark &
& 71.63 & 65.08 & 68.46 & 66.09 & 73.76 & 69.14 & 74.23 & 65.44 & 71.66 & 66.38 & 71.95 & 66.43 \\

 & \checkmark & \checkmark
& 71.54 & 66.09 & 66.23 & 63.72 & 73.84 & 69.93 & 73.63 & 65.59 & 71.81 & 67.39 & 71.41 & 66.54 \\

\checkmark &  & \checkmark
& 71.40 & 65.94 & 68.84 & 66.59 & 73.64 & 69.58 & 73.04 & 64.95 & 71.42 & 66.81 & 71.67 & 66.77 \\

\checkmark & \checkmark &
& 71.20 & 65.49 & 70.34 & 67.76 & 74.38 & 69.01 & 73.89 & 65.24 & 72.16 & 66.91 & 72.39 & 66.88 \\

\midrule

\checkmark & \checkmark & \checkmark
& \cellcolor{softred!35}\textbf{75.13} & \cellcolor{softred!35}\textbf{68.95}
& \cellcolor{softred!35}\textbf{70.80} & \cellcolor{softred!35}\textbf{68.79}
& \cellcolor{softred!35}\textbf{74.63} & \cellcolor{softred!35}\textbf{71.02}
& \cellcolor{softred!35}\textbf{74.99} & \cellcolor{softred!35}\textbf{66.20}
& \cellcolor{softred!35}\textbf{72.51} & \cellcolor{softred!35}\textbf{67.40}
& \cellcolor{softred!35}\textbf{73.61} & \cellcolor{softred!35}\textbf{68.47} \\

\bottomrule
\end{tabular}}
\vskip-3ex
\end{table}

To investigate the effectiveness of the proposed \textbf{FF-CDR} module, as well as the \textbf{TimeELR} and the \textbf{FourierELR} components, we conducted a comprehensive ablation study.
In this experiment, \textbf{SleepDG}~\cite{wang2024generalizable} under $NR=0.6$ symmetric label noise is used as the base baseline (denoted as \textbf{BASE}), upon which different modules are progressively incorporated to evaluate their individual and combined effects.
The ablated methods and their corresponding results are shown in Table~\ref{tab:ablation}.
The component \textbf{TimeELR} yields a moderate improvement over the \textbf{BASE} model, about $0.55\%$ higher in averaged ACC and $2.24\%$ in averaged MF1, indicating that early learning regularization in the time domain can help mitigate overfitting to noisy labels, leading to more stable and reliable predictions.
Similar consequences can also be observed when we solely add \textbf{FourierELR} and \textbf{FF-CDR}.

When replacing \textbf{TimeELR} with \textbf{FourierELR+FF-CDR}, the performance is further improved compared with \textbf{BASE} ($0.77\%$ ACC and $2.48\%$ MF1 for domain-averaged performance), showing that combining frequency-domain early-learning regularization with prediction-level confidence diversity regularization provides stronger robustness against severe label noise than using time-domain ELR alone.
Compared with \textbf{TimeELR}, the addition of \textbf{FF-CDR} leads to a noticeable gain, increasing averaged ACC and MF1 by $0.48\%$ and $0.47\%$, respectively.
Moreover, combining \textbf{TimeELR} with \textbf{FourierELR} achieves higher test-domain averaged performance (ACC: $72.39\%$, and MF1: $66.88\%$), which confirms that frequency-domain regularization provides an effective complement to \textbf{TimeELR} under high-rate label noise. Finally, integrating all components, \ie, our full model, yields the best overall performance, reaching $73.61\%$ averaged ACC and $68.47\%$ averaged MF1.
Compared with \textbf{BASE}, the full model outperforms \textbf{BASE} by $2.97\%$ in averaged ACC and $4.41\%$ in averaged MF1, highlighting the joint collaboration of the three modules.

\subsection{Analysis of Qualitative Results}
\begin{figure}[t]
    \centering

    \begin{subfigure}{1.0\linewidth}
        \centering
        \includegraphics[width=\linewidth]{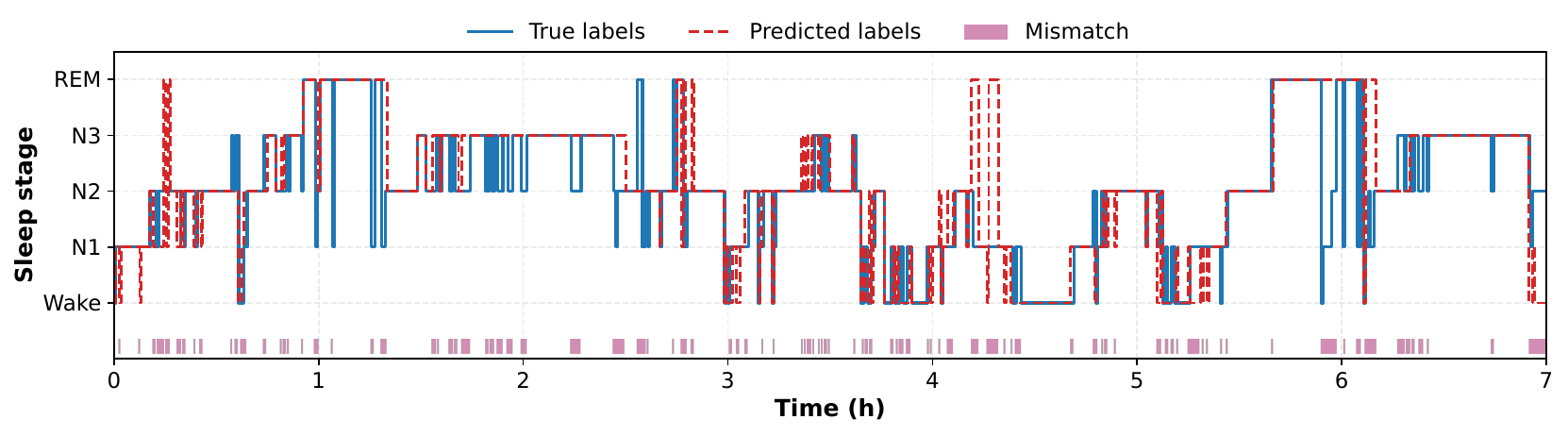}
        \caption{DISC}
        \label{fig:DISC}
    \end{subfigure}

    \begin{subfigure}{1.0\linewidth}
        \centering
        \includegraphics[width=\linewidth]{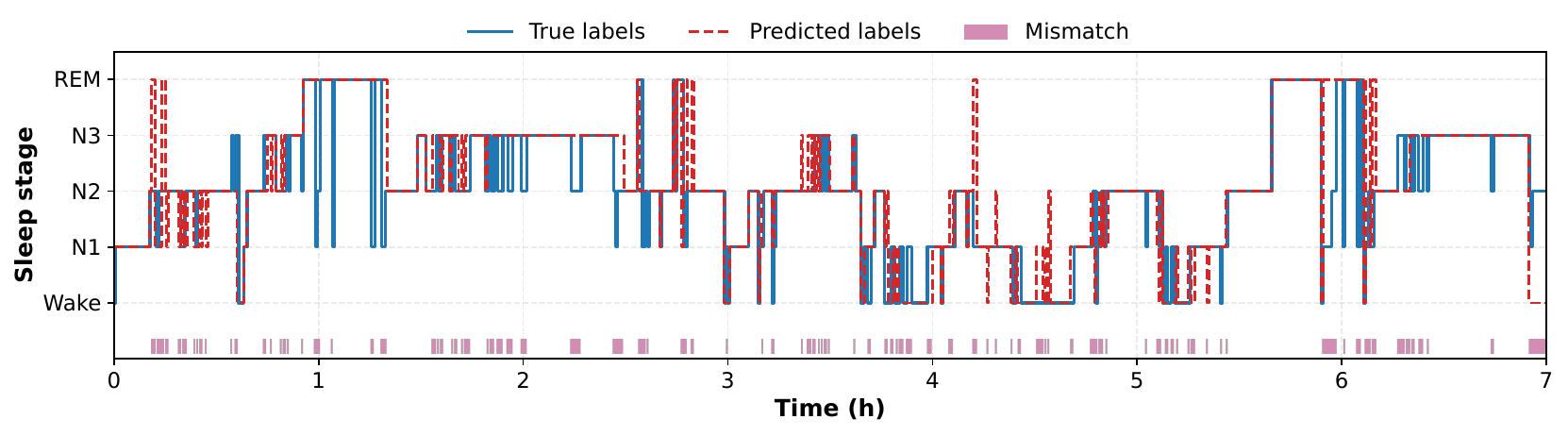}
        \caption{FF-TRUST}
        \label{fig:FF-TRUST}
    \end{subfigure}
    \caption{Comparison of hypnograms generated by the baseline model (a) \textit{DISC} and (b) the proposed \textbf{FF-TRUST} model on Sleep-EDFx under $NR=0.6$ symmetric label noise.
    True sleep stages (blue), predicted stages (red dashed) and mismatches (purple bar at bottom) are visualized over time.}
    \label{fig:sleep_comparison_sleepedfx}
\end{figure}
Figure~\ref{fig:sleep_comparison_sleepedfx} compares hypnograms from the baseline model \textit{DISC}~\cite{li2023disc} and the proposed \textbf{FF-TRUST} model on Sleep-EDFx under $NR=0.6$ symmetric label noise, under which \textbf{FF-TRUST} achieves closer alignment with the ground truth than \textit{DISC}~\cite{li2023disc}, especially in non-REM sleep stages. These results demonstrate that \textbf{FF-TRUST} produces more stable and temporally coherent stage predictions under noisy labels, thereby limiting error propagation across consecutive epochs. By jointly regularizing early-learning signals in both time and frequency domains, and confidence diversity, \textbf{JTF-ELR} anchors optimization to temporally and spectrally stable patterns that are less affected by label corruption, thereby preventing rapid drift toward label noise.
This time-frequency historical consistency constraint dampens gradient oscillations induced by incorrect labels and stabilizes training dynamics, leading to more reliable convergence under severe label noise.

\subsection{Analysis of T-SNE Visualization}
\begin{figure}[t]
  \centering

  \begin{minipage}{0.23\textwidth}
    \centering
    \includegraphics[width=\linewidth]{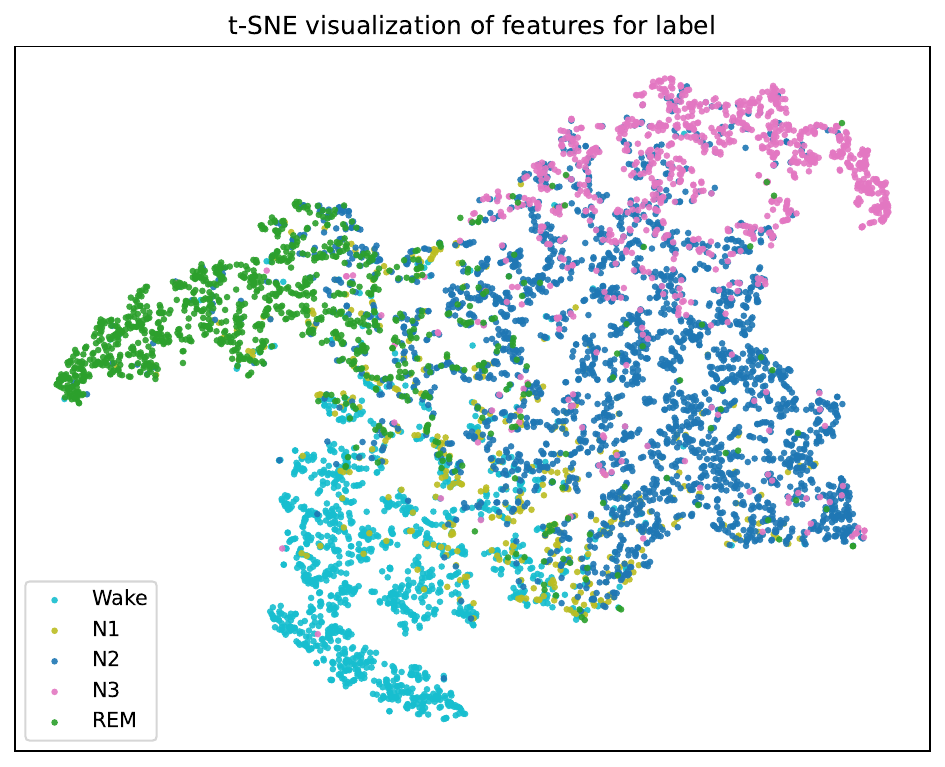}

    {\small (a1)}
  \end{minipage}\hfill
  \begin{minipage}{0.23\textwidth}
    \centering
    \includegraphics[width=\linewidth]{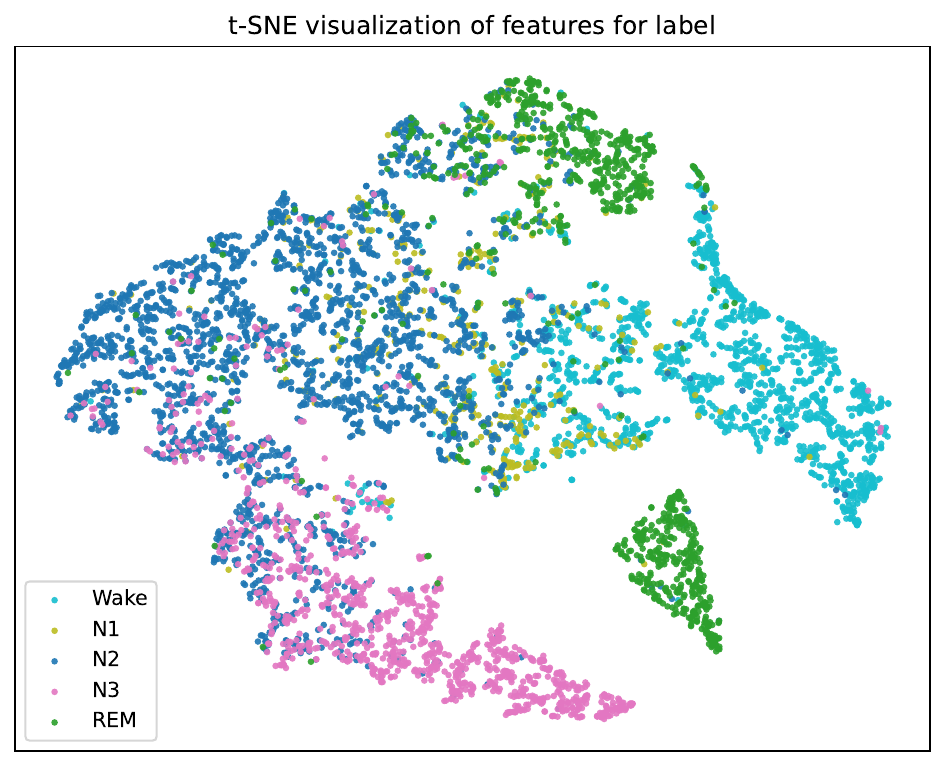}

    {\small (a2)}
  \end{minipage}

  \vspace{-0.2em}
  {\centering\small (a) DISC\par}

  \begin{minipage}{0.23\textwidth}
    \centering
    \includegraphics[width=\linewidth]{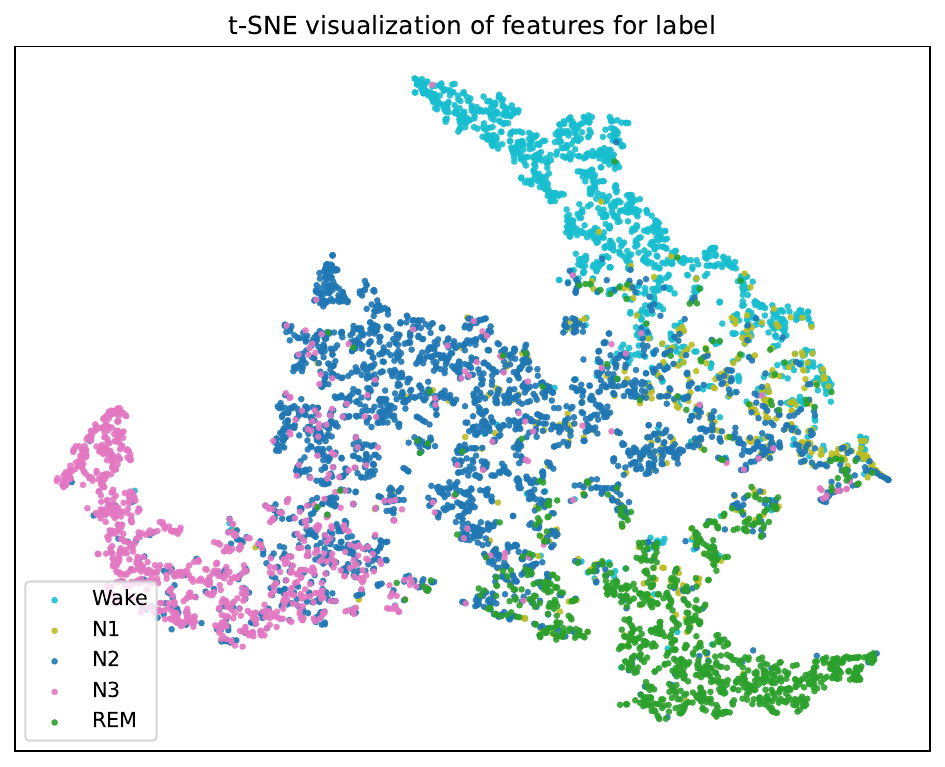}

    {\small (b1)}
  \end{minipage}\hfill
  \begin{minipage}{0.23\textwidth}
    \centering
    \includegraphics[width=\linewidth]{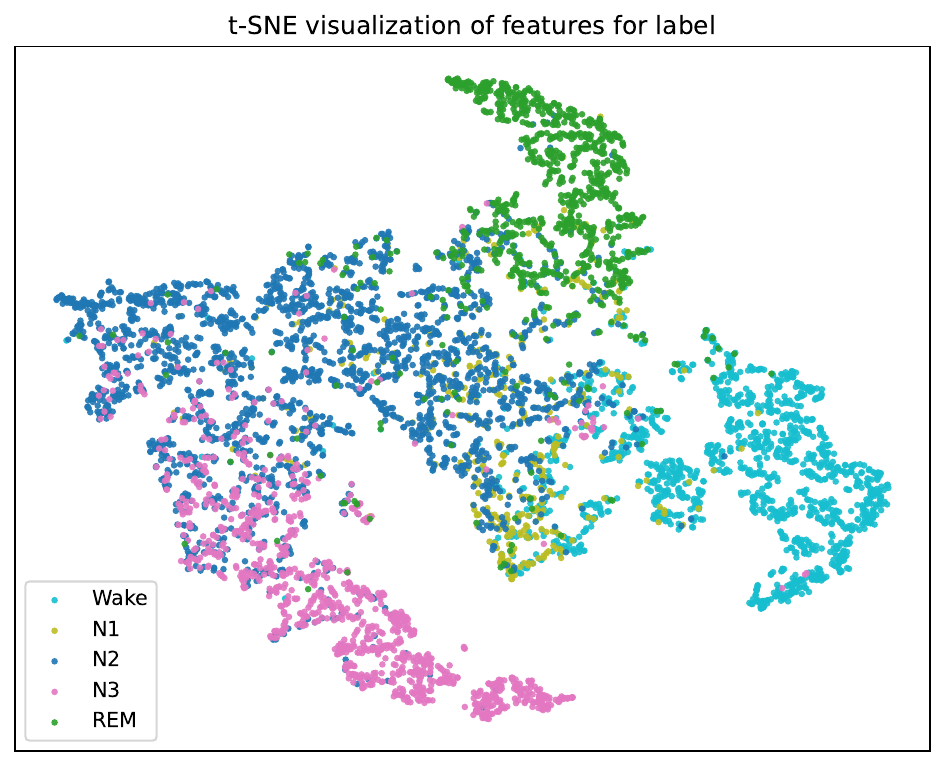}

    {\small (b2)}
  \end{minipage}

  {\centering\small (b) FF-TRUST (Ours)\par}
\vskip-1ex
  \caption{T-SNE visualizations of feature distributions for DISC and \textbf{FF-TRUST} on the HMC and ISRUC datasets.
(a1), (a2) show DISC on HMC and ISRUC, respectively, while (b1), (b2) show \textbf{FF-TRUST}.}
\vskip-3ex
  \label{fig:tsne_DISC_FF-TRUST}
\end{figure}

As shown in Figure~\ref{fig:tsne_DISC_FF-TRUST}, ~\textit{DISC}~\cite{li2023disc} exhibits considerable overlap among different sleep stages and noticeable variations in embedding structures across datasets. In particular, in Figure~\ref{fig:tsne_DISC_FF-TRUST}~\textbf{(a2)}, the REM samples under ~\textit{DISC}~\cite{li2023disc} are fragmented into multiple sub-clusters on the ISRUC dataset, indicating unstable class representations across domains. In contrast, ~\textbf{FF-TRUST} produces more compact and well-separated clusters on both HMC and ISRUC, with highly consistent geometric structures. This suggests that ~\textbf{FF-TRUST} better preserves intrinsic class semantics while mitigating domain-induced feature distortions under label noise perturbations.

\subsection{Analysis of Training Visualization}

\begin{figure}[t!]
    \centering
    \begin{subfigure}[t]{0.48\linewidth}
        \centering
        \includegraphics[width=\linewidth]{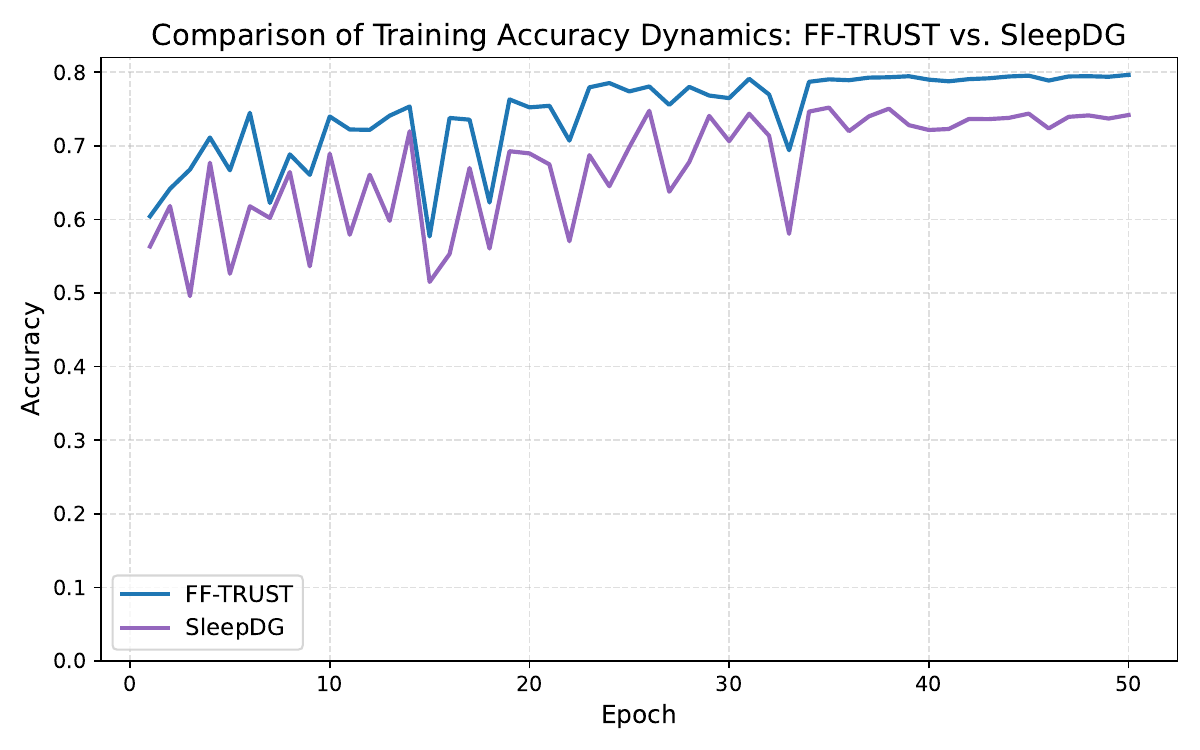}
        \vskip-3ex
        \caption{Training ACC}
        \label{fig:training_acc}
    \end{subfigure}
    \hfill
    \begin{subfigure}[t]{0.48\linewidth}
        \centering
        \includegraphics[width=\linewidth]{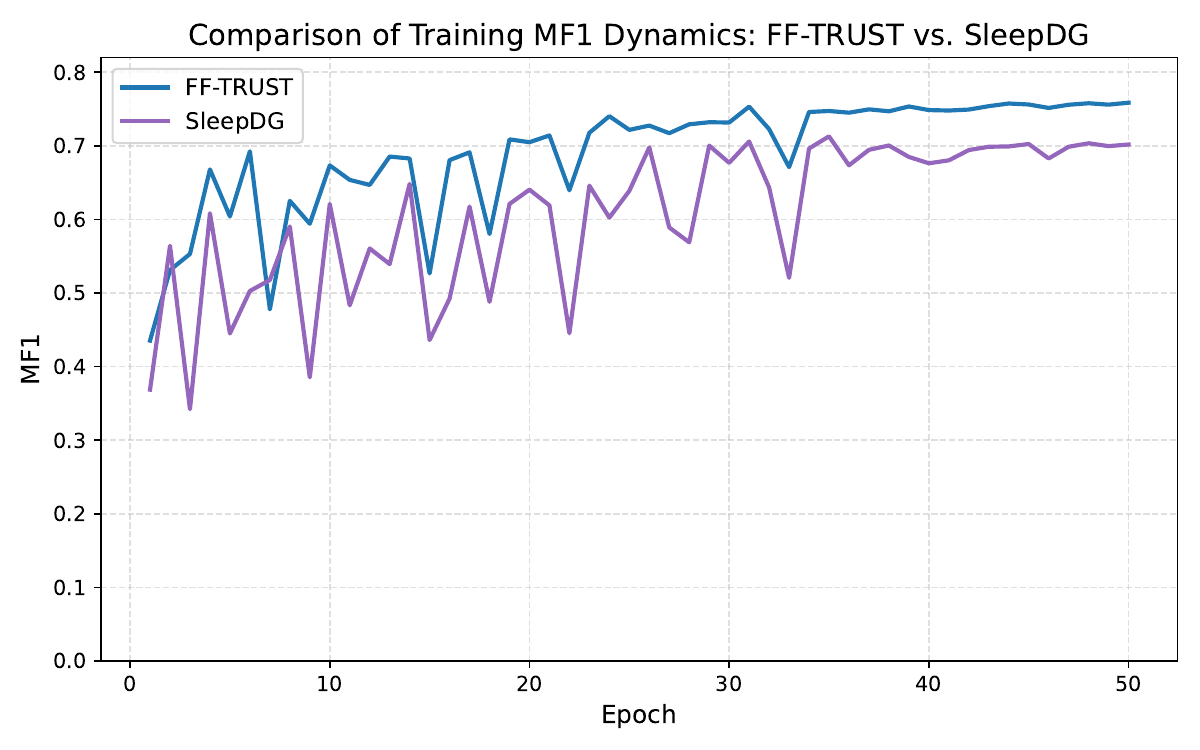}
        \vskip-3ex
        \caption{Training MF1}
        \label{fig:training_mf1}
    \end{subfigure}
    \caption{Comparison of training accuracy (a) and MF1 score (b) over epochs for \textbf{FF-TRUST} and SleepDG on the HMC dataset under $NR=0.6$ asymmetric label noise.}
    \label{fig:training_comparison_HMC}
\end{figure}
Figure~\ref{fig:training_comparison_HMC} visualizes the training accuracy and macro-F1 curves before and after applying the proposed regularization on the HMC dataset under $NR=0.6$ asymmetric label noise. As shown in Figure~\ref{fig:training_comparison_HMC}, \textbf{FF-TRUST} exhibits significantly more stable training
dynamics compared to SleepDG~\cite{wang2024generalizable} under $NR=0.6$ asymmetric label noise.
In terms of training accuracy (Figure~\ref{fig:training_comparison_HMC}~\textbf{(a)}), \textbf{FF-TRUST} demonstrates smoother and more consistent convergence, while SleepDG~\cite{wang2024generalizable} suffers from pronounced
oscillations throughout training.
A similar trend is observed for the MF1 score (Figure~\ref{fig:training_comparison_HMC}~\textbf{(b)}), where
\textbf{FF-TRUST} achieves steadily improving performance and maintains a clear advantage in the later epochs.
These results indicate that the proposed regularization effectively mitigates the adverse impact of asymmetric label noise, leading to more robust and stable optimization behavior. We provide additional details in the Supplementary Material.

\section{Conclusion}

In this work, we introduce \textbf{FF-TRUST} and the first
NL-DGSS benchmark for multimodal sleep staging under noisy
supervision. By integrating temporal-spectral early learning
regularization with forward-free confidence diversity
regularization, FF-TRUST improves cross-domain robustness
without explicit noise-transition estimation. Experiments
on five public datasets demonstrate consistent improvements,
particularly under severe label noise. Despite these results,
the study is limited by synthetic noise assumptions,
magnitude-only spectral representations, and sequence-indexed
EMA buffers that lack independent token-level histories.
Future work will investigate real-world annotation noise,
fine-grained memory tracking, class-prior-aware regularization,
and more comprehensive robustness evaluations.

\balance
\bibliographystyle{IEEEtran}
\bibliography{IEEEabrv,Reference_paper}

\end{document}